\documentclass[12pt]{article}

\usepackage{newtxtext,newtxmath}

\usepackage{graphicx}

\usepackage[letterpaper,margin=1in]{geometry}

\renewenvironment{abstract}
	{\quotation}
	{\endquotation}

\date{}

\makeatletter
\renewcommand{\fnum@figure}{\textbf{Figure \thefigure}}
\renewcommand{\fnum@table}{\textbf{Table \thetable}}
\makeatother

\usepackage{scicite}

\usepackage{url}
\usepackage{amsmath,mathtools}

\usepackage{booktabs}
\usepackage[acronym,nohypertypes={acronym,notation}]{glossaries}
\usepackage{xcolor}
\usepackage{multirow}
\usepackage{multicol}
\usepackage[hidelinks]{hyperref}

\newacronym{sota}{SOTA}{state of the art}
\newacronym{ml}{ML}{mediolateral}
\newacronym{cc}{CC}{craniocaudal}
\newacronym{ap}{AP}{anteroposterior}
\newacronym{par}{PAR}{Paris}
\newacronym{zh}{ZH}{Zurich}
\newacronym{ldn}{LDN}{London}
\newacronym{iv}{IV}{in vivo}
\newacronym{pev}{PEV}{post-euthanasia ventilated}
\newacronym{ev}{EV}{ex vivo}
\newacronym{s}{S}{stereo}
\newacronym{m}{M}{monocular}
\newacronym{vr}{VR}{virtual reality}
\newacronym{phri}{pHRI}{physical human-robot interaction}
\newacronym{roi}{ROI}{region of interest}
\newacronym{cmm}{CMM}{cut, mix \& merge}
\newacronym{iou}{IoU}{intersection over union}
\newacronym{sdr}{SDR}{stereo differentiable rendering}
\newacronym{sdricp}{SDR with in-context prior}{stereo differentiable rendering with in-context prior}
\newacronym{rgb}{RGB}{red-green-blue}
\newacronym{rgbd}{RGB-D}{red-green-blue-depth}
\newacronym{l0}{L0}{link 0}
\newacronym{l1}{L1}{link 1}
\newacronym{l2}{L2}{link 2}
\newacronym{l3}{L3}{link 3}
\newacronym{l4}{L4}{link 4}
\newacronym{l5}{L5}{link 5}
\newacronym{l6}{L6}{link 6}
\newacronym{l7}{L7}{link 7}
\newacronym{ee}{EE}{end effector}
\newacronym{pso}{PSO}{particle swarm optimisation}
\newacronym{cso}{CSO}{camera swarm optimisation}
\newacronym{pnp}{PnP}{Perspective-n-Point}
\newacronym{dlt}{DLT}{direct linear transformation}
\newacronym{hd}{HD}{high definition}
\newacronym{mdr}{MDR}{monocular differentiable rendering}
\newacronym{dof}{DoF}{degrees of freedom}
\newacronym{mse}{MSE}{mean squared error}
\newacronym{ntp}{NTP}{network time protocol}
\newacronym{ct}{CT}{computed tomography}

\usepackage{pifont}
\newcommand{\cmark}{\ding{51}}%
\newcommand{\xmark}{\ding{55}}%

\newcommand{\figref}[1]{Fig.~\ref{#1}}
\newcommand{\secref}[1]{Section "\nameref{#1}"}

\newcommand{\supfigref}[1]{figure~\ref{#1}}
\newcommand{\suptabref}[1]{table~\ref{#1}}

\DeclareMathOperator*{\argmin}{arg\,min}

\def\scititle{
	Localising under the drape: proprioception in the era of distributed surgical robotic system
}
\title{\bfseries \boldmath \scititle}

\author{
	Martin~Huber$^{1}$,
    Nicola~A.~Cavalcanti$^{2}$,
	Ayoob~Davoodi$^{3}$,
	Ruixuan~Li$^{3}$,\and
	Christopher~E.~Mower$^{1,4}$,
    Fabio~Carrillo$^{2}$,
	Christoph~J.~Laux$^{5}$,
	Francois~Teyssere$^{6}$,\and
	Thibault~Chandanson$^{6}$,
    Antoine Harl\'{e}$^{7}$,
	Elie~Saghbiny$^{7}$,
    Mazda Farshad$^{5}$,\and
    Guillaume Morel$^{7}$,
	Emmanuel~Vander~Poorten$^{3}$,
    Philipp F\"urnstahl$^{2}$,\and
	S\'{e}bastien~Ourselin$^{1}$,
	Christos~Bergeles$^{1\ast\dagger}$,
	Tom~Vercauteren$^{1\ast\dagger}$\and
    \small$^{1}$School of Biomedical Engineering \& Imaging Sciences, King's College London, London, United Kingdom\and
    \small$^{2}$Research in Orthopaedic Computer Science, Balgrist University Hospital, Zurich, Switzerland\and
    \small$^{3}$Faculty of Engineering Technology, KU Leuven, Leuven, Belgium\and
    \small$^{4}$Noah's Ark Lab, Huawei, London, United Kingdom\and
    \small$^{5}$University Spine Center Zurich, Balgrist University Hospital, University of Zurich, Zurich, 8008, Switzerland\and
    \small$^{6}$SpineGuard, Paris, France\and
    \small$^{7}$Institut Des Syst\`{e}mes Intelligents et de Robotique, Sorbonne University, Paris, France\and
	\small$^\ast$Corresponding authors. Emails: christos.bergeles@kcl.ac.uk, tom.vercauteren@kcl.ac.uk\and
	\small$^\dagger$These authors contributed equally to this work.
}

\begin{document} 

\maketitle

\begin{abstract} \bfseries \boldmath
Despite their mechanical sophistication, surgical robots remain blind to their surroundings. This lack of spatial awareness causes collisions, system recoveries, and workflow disruptions, issues that will intensify with the introduction of distributed robots with independent interacting arms. Existing tracking systems rely on bulky infrared cameras and reflective markers, providing only limited views of the surgical scene and adding hardware burden in crowded operating rooms. We present a marker-free proprioception method that enables precise localisation of surgical robots under their sterile draping despite associated obstruction of visual cues. Our method solely relies on lightweight stereo-RGB cameras and novel transformer-based deep learning models. It builds on the largest multi-centre spatial robotic surgery dataset to date (1.4M self-annotated images from human cadaveric and preclinical in vivo studies). By tracking the entire robot and surgical scene, rather than individual markers, our approach provides a holistic view robust to occlusions, supporting surgical scene understanding and context-aware control. We demonstrate an example of potential clinical benefits during in vivo breathing compensation with access to tissue dynamics, unobservable under \glsentrylong{sota} tracking, and accurately locate in multi-robot systems for future intelligent interaction. In addition, and compared with existing systems, our method eliminates markers and improves tracking visibility by $25\%$. To our knowledge, this is the first demonstration of marker-free proprioception for fully draped surgical robots, reducing setup complexity, enhancing safety, and paving the way toward modular and autonomous robotic surgery.
\end{abstract}

\section*{One sentence summary}
Marker-free localisation of draped surgical robots 
enables holistic surgical understanding and reveals tissue dynamics.



\noindent
\section*{Introduction}
\label{sec:introduction}
Robotic surgical systems are increasingly replacing conventional approaches across diverse procedures~\cite{sheetz:trends_adoption}. While the advantages robots bring in dexterity, ergonomics, and surgeon well-being are now established~\cite{zarate:ergonomic_analysis,monfared:comparison_ergonomic_risk,stucky:surgeon_symptoms,wells:operating_hurts}, current surgical robots remain effectively unaware of their external environment, or even of each other in distributed multi-arm setups. Most research in surgical automation has primarily focused on tasks inside the patient~\cite{srth,will_your_next_surgeon_be_a_robot}, ignoring that collisions between robotic arms, equipment, and staff continue to pose safety risks~\cite{van:ergonomic_assesment}, disrupt workflows~\cite{wong:improving_ergonomics}, and force system recoveries~\cite{da_vinci_collisions,senhance_collisions}. These risks are expected to intensify with the rise of modular and distributed robotic systems such as Hugo (Medtronic), Versius (CMR Surgical), Senhance (Karl Storz), and Dexter (Distalmotion), which flexibly deploy multiple independent arms around the patient. Current safeguards, such as contact sensors in Hugo, are only last-resort as far as collision detection is concerned. 
Beyond safety, improved spatial awareness can unlock significant efficiency opportunities: automated docking alone, for example, could help shorten operating room setup by as much as 18 minutes of anaesthesia time (within a 146-minute total), which, while routine, carries patient risk and has physiological impact, during robotic adrenalectomy~\cite{feng:robotic_assisted_adrenalectomy,docking_collisions}.

External localisation could overcome these challenges, but current clinically relevant approaches fall short. Infrared tracking systems with reflective markers provide only a partial view of the surgical scene, are easily occluded, require strict calibration, and scale poorly to multi-robot and multi-camera setups~\cite{mmor}. Their cameras are heavy, bulky, costly and difficult to integrate into already crowded operating theatres. Marker-free localisation, while effective in industrial robotics where robots are clearly observable and their bases are fixed~\cite{easyhec, ctrnet, automatic_hand_eye}, is fundamentally incompatible with surgical practice. There, sterility protocols mandate that robots be draped before entering the operative field, with drapes obscuring visual cues that would be picked up by conventional tracking systems 
(\figref{fig:results.overview}). Moreover, clinical workflow requires robot placement to be adapted to each patient’s individual needs, even during surgery, making pre-procedural calibration approaches inadequate. These constraints motivate the development of marker-free localisation approaches tailored to surgical robots.

This work is the first to introduce a marker-free localisation framework tailored to draped surgical robots (\figref{fig:results.method_overview}).
Our proposed method, detailed in the \nameref{sec:methods} section, uses a lightweight low-cost stereo \gls{rgb} camera in combination with \glsentrylong{sdr} to estimate a draped robot pose. 
\Glsentrylong{sdr} aligns virtual/rendered robot models to silhouette segmentations inferred by a deep learning model applied to observed images.
Silhouette segmentation under severe occlusion from tools, cables, staff, and sterile draping, is achieved thanks to our multi-centre spatial robotic surgery dataset and the developed transformer-based deep learning models. Our dataset is the largest to date, comprising 1.43 million images, self-annotated via precise robot localisations, across in vivo and ex vivo porcine studies and human cadaveric procedures. 
Segmentation accuracy is further refined by our novel \glsentrylong{cmm} augmentation, which simulates realistic multi-robot setups, and our proposed refinement scheme: \glsentrylong{sdricp}. 

We evaluate our framework in robotic spine surgery, specifically pedicle screw placement, as an exemplar clinical benchmark. This procedure demands submillimetre precision for screw placement and is further complicated by continuous spine motion due to patient breathing~\cite{clinical::campbell2021clinical,clinical::mckenzie2021robotics,clinical::perfetti2022robotic}. Unlike minimally invasive laparoscopic procedures, tracking is already part of routine clinical practice in robotic spine surgery. This allows for a direct comparison between our approach and existing methods, which rely on invasive fiducial markers that are drilled into the vertebral body. Markers are often distant from the surgical field, which can necessitate a holistic view of the spine to accurately track its motion. The combination of high accuracy, robustness to motion, and limitations of current tracking practices makes pedicle screw surgery a particularly challenging yet representative benchmark for evaluating marker-free localisation in surgical robotics.

We validate our approach under both preclinical and clinical workflow conditions during porcine in vivo and human cadaveric spine surgery. In our preclinical studies, \glsentrylong{sdr} enabled marker-free localisation of draped surgical robots and supported breathing-motion compensation during robotic pedicle screw placement. Building on this foundation, we successfully transitioned to clinical workflow conditions with localisation in multi-robot setups via \glsentrylong{sdricp}, paving the way for future multi-robot intelligent coordination. Compared to infrared tracking, our method reduced hardware burden more than tenfold, improved surgical-site visibility by $25\%$, and revealed tissue motions invisible to conventional systems. Critically, it achieved sub-percent localisation accuracy at the robot base ($<0.16\%$ / $<4\,\text{mm}$ error at $2.6\,\text{m}$ distance) and outperformed marker-based methods at the surgical tool centre even with unknown tool.

More broadly, the principles of \glsentrylong{sdr}, in-context priors, and occlusion-invariant segmentation can extend to multi-robot platforms that must robustly operate in cluttered and occluded environments beyond surgery.

\section*{Results}
\label{sec:results}
To demonstrate the relevance of our work, we initially located undraped surgical robots inside the sterile zone and subsequently draped them in place (\figref{fig:results.overview}B). This draping procedure facilitated a unique dual opportunity to:
(i) assess the clinical feasibility and relevance of \hyperref[sec:results.marker_free_surgical_robot_localisation_in_preclinical_workflow]{\emph{marker-free surgical robot localisation in preclinical workflow}} using the proposed \glsentrylong{sdr} approach (\figref{fig:results.method_overview});
and (ii) facilitate the gradual transition of the \glsentrylong{sdr}-based localisation strategy from preclinical experiments to \hyperref[sec:results.localising_draped_surgical_robots_inside_the_sterile_zone]{\emph{localising draped surgical robots inside the sterile zone}} for clinically realistic workflow conditions (\figref{fig:results.transition}).

\subsection*{Multi-centre studies design and overview}
\label{sec:results.multi_centre_studies_design_and_overview}
The studies presented in this work were carried out in four centres in Europe: two in \glsentrylong{ldn}, one in \glsentrylong{zh}, and one in \glsentrylong{par} (\suptabref{tab:results.dataset}, \figref{fig:results.transition}). In all studies, surgical robots were initially localised without draping, then draped in place. For preclinical experiments (deviating from clinical protocol), this demanded draping inside the sterile zone.


The \glsentrylong{ldn} mock spine surgery benchmark dataset (\supfigref{fig:results.sie_benchmark_data}) was acquired to replicate and quantify the draping procedure inside the sterile zone. There, the surgical robot (LBR Med 7, KUKA AG, Augsburg, Germany; $\pm0.15\,\text{mm}$ repeatability) was equipped with an AprilTag for marker-based reference localisation. A stereo-\gls{rgb} camera (ZED 2i, Stereolabs Inc., San Francisco, CA, USA) was mounted bedside on a standard laparoscopic arm (Articulated Stand straight, Karl Storz SE \& Co. KG, Tuttlingen, Germany). The surgical robot was moved into fifteen configurations within the spinal workspace, then draped, and finally the configurations were replayed. This procedure was repeated for three camera poses (left lateral thoracic, sacral, and thigh levels). Two additional datasets (\glsentrylong{ldn} I and II) were acquired for ground-truth generation (supplementary).

Preclinical studies were conducted in \glsentrylong{zh} and \glsentrylong{par}. The \glsentrylong{zh} studies replicated the mock spine surgery benchmark setup (\figref{fig:results.overview}B, \supfigref{fig:results.sie_benchmark_data}) and also incorporated clinical equipment for automated robotic pedicle screw placement surgery, detailed in \secref{sec:results.preclinical_robotic_pedicle_screw_placement_surgery_setups}. The \glsentrylong{zh} studies progressively worked toward in vivo breathing compensation, conducting procedures on human specimen (Global Anatomix (GAX), Miami, FL, USA; $n=1$), then porcine \glsentrylong{ev} ($n=1$), and finally porcine \glsentrylong{iv} ($n=1$) models. The human specimen extended from the hips to the shoulders (\figref{fig:results.method_overview}, left). Datasets were collected concurrently with clinical relevance evaluation. The preclinical \glsentrylong{par} porcine \glsentrylong{iv} ($n=1$) study stem from an earlier project iteration and vision was not yet incorporated into breathing compensation. However, two surgical robots were deployed, providing a benchmark for localising under the drape in a multi-robot setup.

\subsection*{Preclinical robotic pedicle screw placement surgery setups}
\label{sec:results.preclinical_robotic_pedicle_screw_placement_surgery_setups}
The preclinical surgical setup (deployed across subjects in \glsentrylong{zh} and \glsentrylong{par}) is shown in \figref{fig:results.overview}A. The surgical robot (LBR Med 7, KUKA AG, Augsburg, Germany; two for the \glsentrylong{par} study, see \figref{fig:results.segmentations}) was rigidly clamped to a height-adjustable operating table equipped with standard padding and limb extensions. A custom, motor-driven drill (EC 60 flat, Maxon Motor AG, Sachseln, Switzerland) equipped with a tool-mounted infrared marker was connected to the robot via a tool changer (G-SHW063, GRIP GmbH, Dortmund, Germany), with a Nano25 force/torque sensor (ATI Industrial Automation Inc., Apex, NC, USA) mounted in between. The sensor measured drill-applied forces (\figref{fig:results.overview}B). The drill bit (SpineGuard SA, Vincennes, France) had a diameter of $3.5\,\text{mm}$ and an exposed length of $175\,\text{mm}$. The same surgical robot and drill system were used in all experiments.

Stereo-\gls{rgb} cameras (ZED 2i, Stereolabs Inc., San Francisco, CA, USA; $166\,\text{g}$, $172\,\text{mm}$ each) were used for marker-free localisation. For clinical comparison, a stereo-infrared camera (fusionTrack 500, Altracsys LLC, Puidoux, Switzerland; $2160\,\text{g}$, $528\,\text{mm}$) was placed within the operating theatre (\figref{fig:results.overview}A). The stereo-\gls{rgb} cameras were therefore each approximately $13\times$ lighter and $3\times$ smaller than the infrared tracking system, offering a compact and flexible alternative. In the all \glsentrylong{zh} studies, two stereo-\gls{rgb} were used. One was mounted bedside on a standard laparoscopic arm (Articulated Stand straight, Karl Storz SE \& Co. KG, Tuttlingen, Germany), and a second unit was ceiling-mounted on a custom mechanical arm (\figref{fig:results.method_overview}, left). In the \glsentrylong{par} porcine in vivo study, one stereo-\gls{rgb} was wall-mounted.

\subsection*{Preclinical surgical workflow}
\label{sec:results.preclinical_surgical_workflow}
In preparation for the surgical procedures, spine \acrshort{ct} scans were acquired to enable pedicle screw trajectory planning. Vertebrae were segmented and converted into surface meshes using 3D volume processing software (Mimics Innovation Suite, Materialise NV, Leuven, Belgium). Following the procedure described in~\cite{clinical::massalimova2023automatic}, bilateral pedicle trajectories from T10 to L5 were planned using an institutional surgical planning platform (CASPA, Zurich, Switzerland).

In this controlled preclinical workflow, with the robot rigidly attached, we performed the proposed \glsentrylong{sdr}–based localisation (\figref{fig:results.method_overview}) of the undraped surgical robot inside the sterile zone using stereo-\gls{rgb} cameras, see \secref{sec:results.marker_free_surgical_robot_localisation_in_preclinical_workflow}. For a clinical gold standard reference, we performed marker-based localisation with the clinical stereo-infrared camera using \glsentrylong{pnp} and the tool infrared marker (\figref{fig:results.overview}A). Following the localisation, the system was draped in place, and all subjects were placed prone on the operating table with vertebral levels identified to ensure consistent positioning. Vertebral level identification was performed using a handheld linear ultrasound probe (ML6-15-D, GE Healthcare, Chicago, IL, USA).

All experiments were performed under consistent preclinical conditions replicating an operating-room environment (\figref{fig:results.overview}). The dorsal region was prepared and draped in a sterile fashion. For surgical access, all specimens underwent a Wiltse approach on the left and a standard midline approach on the right~\cite{clinical::wiltse1973paraspinal,clinical::street2016comparison}. The exposed spine was then registered to the surgical robot using landmark-based registration~\cite{clinical::Suter:CRAS:2022}. The robot was then aligned to the planned trajectory and drilling proceeded at $3000\,\text{rpm}$. Drilling stopped at the planned depth.

\subsection*{Preclinical porcine in vivo surgical workflow}
\label{sec:results.preclinical_porcine_in_vivo_surgical_workflow}
The \glsentrylong{zh} and \glsentrylong{par} porcine in vivo studies followed the preclinical surgical workflow. Ethics approvals were granted by the respective Swiss and French institutions (see ethics approvals in \secref{sec:acknowledgments}). Veterinary anaesthesiologists supervised intravenous anaesthesia, mechanical ventilation, and full perioperative monitoring.

Stemming from an earlier project stage, breathing compensation was not visually assisted in the \glsentrylong{par} study. For the \glsentrylong{zh} study, clinically certified markers were applied to monitor breathing motion to support the drilling. Using markers here adhered to the highest clinical standards, and supported robust ground-truth respiratory estimates. For the clinical stereo-infrared tracking system, a fiducial infrared marker was mounted on a percutaneous $2.5\,\text{mm}$ K-wire placed through a stab incision through the left lamina of T7 and advanced into the vertebral body to achieve rigid bony purchase. Placement occurred within the sterile field and outside the drapes to preserve line-of-sight for optical tracking. To overcome the sparse and invasive tracking facilitated by tracking a single infrared marker, clinically certified AprilTags (VisiMARKERs, Clear Guide Medical, Baltimore, MD, USA) were affixed to the anatomy for the bedside-mounted stereo-\gls{rgb} camera (\figref{fig:results.overview}B). Placing the adhesive AprilTags took less than thirty seconds. One additional AprilTag (ID 5) was attached immediately to the distal fiducial infrared marker for reference. The \glsentrylong{par} and \glsentrylong{zh} subjects were humanely euthanised under deep anaesthesia after more than five hours of surgery. In the \glsentrylong{zh} study, surgery continued \glsentrylong{pev}.

\subsection*{Benchmarking marker-free surgical robot localisation}
\label{sec:results.benchmarking_marker_free_surgical_robot_localisation}
We first quantified the localisation accuracy of \glsentrylong{sdr} (see \figref{fig:results.method_overview}, \secref{sec:methods.stereo_differentiable_rendering}) on the undraped system on the \glsentrylong{ldn} mock spine surgery benchmark dataset  (\supfigref{fig:results.sie_benchmark_data}). In the benchmark, the robot was often only partially visible.

We performed a Monte Carlo cross-validation with five randomly sampled robot configuration sets per camera pose for $3$, $6$, $9$, and $12$ robot configurations, each, totalling $60$ unique localisations. We evaluated the localisation accuracy on the complementary robot configurations. Localisation results are summarised in \figref{fig:results.localisation_errors_benchmark}. An initial camera pose estimate was obtained using the \gls{rgb}-depth-based Hydra algorithm~\cite{hydra}. Localisation was then refined using our proposed \glsentrylong{sdr} algorithm (\figref{fig:results.method_overview}). In this phase of the work, no pre-trained robot segmentation model was available,
we thus relied on the foundation segmentation model SAM~2~\cite{sam2} with manual prompting for silhouette extraction.
We further compared against a \glsentrylong{pnp}-based approach (clinial gold standard)~\cite{hydra}, which assumed exact knowledge of the tool marker location with respect to the robot (known tool). Note that the randomised sets were the same for every method. It can be seen (\figref{fig:results.localisation_errors_benchmark}) that the \glsentrylong{pnp}-based approach performed best with [0.9, 0.56, 0.61, 0.48] mm median deviation from the surgical tool centre for [3, 6, 9, 12] robot configurations, respectively. The \glsentrylong{sdr}-based approach (using SAM~2) scored slightly worse with [1.17, 1.06, 0.94, 0.9] mm median deviation. However, the benchmark dataset was biased towards the \glsentrylong{pnp}-based approach for two reasons. First, only robot configurations where the marker was visible were considered. Second, the reference AprilTag was not part of the 3D robot model for \glsentrylong{sdr}. The reported localisation error for \glsentrylong{sdr} is thus expected to be an upper bound.

\subsection*{Marker-free surgical robot localisation in preclinical workflow}
\label{sec:results.marker_free_surgical_robot_localisation_in_preclinical_workflow}
We next deployed the marker-free \glsentrylong{sdr}-based localisation under the \hyperref[sec:results.preclinical_surgical_workflow]{\emph{preclinical surgical workflow}} across all \glsentrylong{zh} and \glsentrylong{par} studies.
Since the workflow remained the same across studies and subjects, and to emphasise the clinical relevance, we here focus on the porcine in vivo study. 

Quantitative results were provided above. Qualitative results for the localisation accuracy are shown in \figref{fig:results.overview}B, where the green wireframe render coincides with the surgical robot underneath the sterile draping. During the in vivo study, an initial camera pose estimate was obtained by hand-guiding~\cite{lbr_stack} the undraped robot into 20 unique configurations (now unconstrained by marker-visibility), followed by the marker-free localisation procedure from \secref{sec:results.benchmarking_marker_free_surgical_robot_localisation}. Localisation was executed in parallel for the bedside- and ceiling-mounted stereo-\gls{rgb} cameras. Both cameras were co-registered to the surgical robot  (\figref{fig:results.method_overview}, point cloud and render). Localisation during the in vivo study was achieved in under five minutes, therefore having minimal impact on the duration of the surgical procedure. Under these preclinical conditions, the surgical robot was then successfully draped in place without changing its location. We thus demonstrated potential clinical feasibility of marker-free localisation via \glsentrylong{sdr}, subject to \hyperref[sec:results.drape_and_occlusion_invariant_segmentation_of_surgical_robots]{\emph{drape- and occlusion-invariant segmentation of surgical robots}} for clinically realistic workflow conditions.

\subsection*{Proprioceptive breathing motion estimation in the robot reference frame}
\label{sec:results.proprioceptive_breathing_motion_estimation_in_the_robot_reference_frame}
Beyond feasibility, we next studied the clinical relevance of our marker-free localisation approach during the \glsentrylong{zh} porcine in vivo study. The proposed marker-free \glsentrylong{sdr}, and the infrared marker-based \glsentrylong{pnp} localisation approaches, both allowed to map the subject breathing motion into the robot base frame, displayed for the bedside-mounted stereo-\gls{rgb} camera in \figref{fig:results.overview}B (AprilTags, green wireframe). Breathing-motion tracking used the clinically certified AprilTags (non-invasive, distributed) and the infrared marker (invasive, single reference) introduced in \secref{sec:results.preclinical_porcine_in_vivo_surgical_workflow}. Respiratory motion in the stereo-\gls{rgb} camera frame was estimated via \hyperref[sec:methods.stereo_marker_tracking]{\emph{stereo marker tracking}}, which remains fully compatible with future marker-free tracking. The distal fiducial infrared marker was tracked using proprietary software from the stereo-infrared camera vendor.

Estimated breathing trajectories at the fiducial infrared marker and the reference AprilTag (ID 5) closely matched, as qualitatively shown for a drilling case in \figref{fig:results.breathing} (blue and green curves, respectively). Breathing-motion estimates were well within submillimetre accuracy and clinically acceptable for pedicle screw placement surgery, suggesting clinical parity. Additional measurements comparing motion to an AprilTag clamped directly to the T15 spinous process (\figref{fig:results.overview}A) showed that the proximal AprilTag (ID 10; $24.8 \pm 0.1\,\text{cm}$ from drilling site) provided more physiologically representative motion estimates, whereas the infrared marker ($32.9 \pm 0.1\,\text{cm}$ from drilling site) showed inconsistent motion (\suptabref{tab:results.breathing_amplitudes}). Furthermore, because the surgeon intermittently obstructed the line of sight, the fiducial infrared marker was visible only $77.3\%$ of the time. By comparison, despite variable lighting conditions, the AprilTag was successfully triangulated $96.6\%$ of the time, yielding a $25\%$ relative increase in visibility (detailed evaluations in supplementary).

These results demonstrate that the distributed AprilTags provided a holistic view of the respiratory motion that was both physiologically accurate and more robust than the single infrared marker.

\subsection*{Breathing-compensated drilling in robotic porcine in vivo spine surgery}
\label{sec:results.breathing_compensated_drilling_in_robotic_porcine_in_vivo_spine_surgery}
We next studied whether respiratory estimates could support informed and autonomous motion compensation during the \glsentrylong{zh} porcine in vivo study. We incorporated the visually obtained proprioceptive breathing motion estimates into a contact force controller for \hyperref[sec:methods.sensorimotor_breathing_compensation]{\emph{sensorimotor breathing compensation}}, see force/torque in \figref{fig:results.overview}B. Drilling sequences transitioned from breathing-compensated drilling (pre- and post-contact), to drill stop, and finally drill retraction (\figref{fig:results.breathing}). Breathing was not compensated during the drill stop or retraction phases.

The transition from pre- to post-contact was characterised by a steady force ramp-up (\figref{fig:results.breathing}, light blue curve), consistent with the improved breathing amplitude capture at the proximal AprilTag (previous section, \suptabref{tab:results.breathing_amplitudes}). During drilling, a force of $14.78 \pm 0.76\,\text{N}$ was applied to the spine, reasonably within the target force of $15\,\text{N}$.
The spine-applied target force induced tissue displacement from the steady state throughout the breathing-compensated drilling phase and as the drill advanced into the pedicle (\figref{fig:results.breathing} and \suptabref{tab:results.breathing_dynamics}, quadratic fits). Along the \glsentrylong{ml} and \glsentrylong{ap} axes, the displacement direction aligned with the drill tip velocity across all tracking approaches. However, the proximal AprilTag reported a $325\%$ faster anterior displacement ($-0.374 \pm 0.026\,\text{mm/s}$) than the fiducial infrared marker ($-0.088 \pm 0.005\,\text{mm/s}$), suggesting a more pronounced deformation at the drilling site itself.
Along the \glsentrylong{cc} axis, the infrared system detected displacement opposite to the kinematically tracked drilling direction, whereas stereo-\gls{rgb}-based tracking produced physiologically consistent cranial displacement within a standard deviation. Because only respiratory amplitudes were compensated in this study, the resulting tissue displacement emerged as an unforeseen observation, and the additional deformation information was not at that stage exploited for trajectory adaptation.

On drill stop, the animal was maximally displaced and breathing compensation was deactivated. The largest displacement occurred along the \glsentrylong{ml} axis ($3.384 \pm 0.129\,\text{mm}$ at the proximal AprilTag), closely matching the infrared marker. Consistent with earlier breathing amplitude observations, the fiducial infrared marker exhibited a physiologically implausible displacement opposite the drilling direction along the \glsentrylong{cc} axis. Along the \glsentrylong{ap} axis, infrared tracking reported a submillimetre displacement of $-0.529 \pm 0.056\,\text{mm}$, whereas the proximal AprilTag measured a $250\%$ greater displacement of $-1.850 \pm 0.225\,\text{mm}$, indicating that marker placement strongly influences motion representation.

With breathing compensation disabled, a force of $16.7 \pm 1.8\,\text{N}$ was induced at the drill tip. This coincided with posterior breathing amplitude suppression of $1.19 \pm 0.10\,\text{mm}$ at the proximal AprilTag and $0.29 \pm 0.08\,\text{mm}$ at the infrared marker. These correspond to relative reductions of $28 \pm 5\%$ and $10 \pm 3\%$, respectively, which is $310\%$ more at the AprilTag. Notably, along the \glsentrylong{ml} axis, the breathing amplitude even reversed direction from left to right (\figref{fig:results.breathing}). Upon drill retraction, the displaced spine gradually relaxed back to steady state. 

These findings confirm that stereo-RGB-based tracking provides a more complete and physiologically representative view of spinal motion than sparse infrared tracking and demonstrate successful proprioceptive autonomous action based on \hyperref[sec:results.marker_free_surgical_robot_localisation_in_preclinical_workflow]{\emph{marker-free surgical robot localisation in preclinical workflow}}.



\subsection*{Drape- and occlusion-invariant segmentation of surgical robots}
\label{sec:results.drape_and_occlusion_invariant_segmentation_of_surgical_robots}
Accurate localisation of draped robots required drape- and occlusion-invariant segmentation of surgical robots, which deviates substantially from traditional segmentation tasks. The difficulty of segmenting a surgical robot that is largely invisible is further illustrated in \figref{fig:results.segmentations}A, where draping accumulated and turned opaque, while cables and staff frequently obscured the surgical robot. 

To address this challenge, we generated several large spatial robotic surgery datasets: $1.43$ million fully self-annotated images across porcine \glsentrylong{ev} / \glsentrylong{iv} / \glsentrylong{pev}, and human GAX specimen surgeries (\suptabref{tab:results.dataset}). 
Self-annotation was achieved by  \hyperref[sec:results.marker_free_surgical_robot_localisation_in_preclinical_workflow]{\emph{marker-free surgical robot localisation in preclinical workflow}} combined with synchronised joint positions to kinematically generate accurate ground-truth segmentations over time for all procedures (\figref{fig:results.transition}, \figref{fig:results.segmentations}A). Refer to \suptabref{tab:results.dataset} and supplementary for details. The \glsentrylong{par} porcine in vivo dataset (\figref{fig:results.segmentations}A, \suptabref{tab:results.dataset}) was selected for testing, due to its challenging draping conditions, unique operating theatre setup, and inclusion of multiple robots (all training datasets contained a single robot). To accelerate architecture search, we subsampled the training dataset of $1.2$ million image-render pairs into two smaller datasets: medium (47,600 pairs, $4\%$) and large (95,364 pairs, $8\%$), where only frames with non-negligible joint motion were considered for maximum data diversity.

When trained on the medium dataset, the transformer-type network with MIT-B3-based encoder consistently outperformed the convolution-type network in \gls{iou}, refer \suptabref{tab:results.iou}. The \glsentrylong{cmm} augmentation (\figref{fig:results.transition}) facilitated generalisation to multi-robot setups ($0.64\,/\,0.69$ \gls{iou} without / with \glsentrylong{cmm} augmentation; \figref{fig:results.segmentations}B vs. C). Thus, with only one-fifth the size of the foundation model SAM~2, and not requiring user interaction, the MIT-B3-based model achieved significantly better accuracy  ($0.69 \pm 0.07$ vs. $0.60 \pm 0.03$ \gls{iou}). The highest accuracy ($0.73$ \gls{iou}) was then achieved by a MIT-B5-based model when trained on the large dataset (\figref{fig:results.segmentations}E). A zero in-context prior, used for \hyperref[sec:methods.stereo_differentiable_rendering_with_render_prior]{\emph{stereo differentiable rendering with in-context prior}}, left \gls{iou} unchanged.

\subsection*{Localising draped surgical robots inside the sterile zone}
\label{sec:results.localising_draped_surgical_robots_inside_the_sterile_zone}
Moving into the sterile zone under clinically realistic workflow conditions, in this section we evaluated the transferability of the proposed \glsentrylong{sdr} algorithm, using \hyperref[sec:results.drape_and_occlusion_invariant_segmentation_of_surgical_robots]{\emph{drape- and occlusion-invariant segmentation of surgical robots}} for the robot segmentor introduced in \figref{fig:results.method_overview}. We utilised the MIT-B5-based three / four channel models (without / with in-context prior, \suptabref{tab:results.iou}), where the in-context prior served in an iterative refinement scheme (see \figref{fig:results.render_prior}, \secref{sec:methods.stereo_differentiable_rendering_with_render_prior}). 

As already in \hyperref[sec:results.marker_free_surgical_robot_localisation_in_preclinical_workflow]{\emph{marker-free surgical robot localisation in preclinical workflow}}, we initially investigated tool centre localisation accuracy via Monte Carlo cross-validation on the \glsentrylong{ldn} mock spine surgery benchmark dataset (\supfigref{fig:results.sie_benchmark_data}) with results provided in \figref{fig:results.localisation_errors_benchmark}, then moved toward evaluating on the \glsentrylong{par} multi-robot in vivo test dataset (\figref{fig:results.segmentations}) with results in  \figref{fig:results.localisation_errors_clinical}. For all draped localisation experiments, we found an initial camera pose estimate through the proposed \glsentrylong{cso} algorithm (\secref{sec:methods.stereo_differentiable_rendering}). An exemplary initial camera pose estimate is shown in \figref{fig:results.render_prior} (in-context prior). The \glsentrylong{sdricp} method was run for a total of $200$ iterations, and the segmentation got updated every $50$th iteration. In addition to the clinical gold standard \glsentrylong{pnp}-based approach (known tool), we evaluated our method against the best marker-based approach under an unknown tool~\cite{hec_shah}. 

On the draped \glsentrylong{ldn} mock spine surgery benchmark data, SAM~2 guided \glsentrylong{sdr} prove infeasible, with median errors of [18, 16, 18, 19] mm, well in the centimetre regime, hence omitted in \figref{fig:results.localisation_errors_benchmark}. The \glsentrylong{pnp} method performed best with [0.9, 0.56, 0.61, 0.48] mm, but it required exact knowledge of the marker location with respect to the robot. This assumption may not hold in complex clinical scenarios with varying tools under modular integrated systems, and does rely on markers. In this challenging draped scenario, the in-context prior prove beneficial. We found localisation accuracy improvements from [4.79, 2.41, 2.72, 1.63] mm to [2.87, 2.17, 1.78, 1.33] mm, corresponding to average relative improvements of $26\%$. The in-context prior further reduced error ranges from [5.63, 3.02, 2.48, 2.19] mm to [3.11, 1.67, 1.47, 1.05] mm, corresponding to an average error range reduction of $46\%$. When comparing \glsentrylong{sdricp} on the undraped vs. the draped scenario, an added submillimetre localisation error of [1.04, 0.86, 0.57, 0.36] mm was found, suggesting that the draping-induced error was effectively mitigated. Finally, it was found that \glsentrylong{sdricp} was [1.36, 0.96, 0.73, 0.55] mm more accurate than the marker-based localisation under unknown tool, providing a relative improvement of $30\%$.



On the in vivo multi-robot \glsentrylong{par} test dataset, we evaluated the proposed \glsentrylong{sdr} methods for repeatability, i.e. undraped localisation was compared against draped localisation results. Therefore, we selected $36$ unique draped robot configurations from the \glsentrylong{par} test dataset. Due to workspace limitations during the in vivo pedicle screw placement surgery, we retrospectively selected those $36$ configurations for improved variability in joint space by K-means clustering. The $36$ robot configurations, as well as corresponding images, were then split into $4$ unique sets for Monte Carlo cross-validation with $9$ calibration samples each.

\figref{fig:results.localisation_errors_clinical} presents the repeatability that was achieved by the proposed \glsentrylong{sdr} without / with in-context prior. The repeatability was evaluated on the respective complementing samples for the Monte Carlo fit subsets along the kinematic chain, including \glsentrylong{l0} up to the \glsentrylong{ee}. It was found that the median deviation from the undraped localisation was $0.96\,/\,0.39\,\text{cm}$ at \glsentrylong{l0}, and $2.34\,/\,1.66\,\text{cm}$ at the \glsentrylong{ee}, corresponding to relative improvements of $59\%$ and $29\%$. Since the camera was wall-mounted and distanced $2.6\,\text{m}$ from the robot base, these errors corresponded to median deviations of $< 0.38\,/\,0.16\%$ at \glsentrylong{l0}, and $< 0.91\,/\,0.65\%$ at the \glsentrylong{ee}, respectively.

\section*{Discussion}
\label{sec:discussion}
Our work was borne out of the need to provide a framework for the holistic support of robotic surgeons and staff in the operating theatre today, moving away from automation of isolated tasks. 
Supporting staff required several methodological contributions, most notably the development of \glsentrylong{sdr} for localisation under the drape. We further generated and self-annotated multi-centre spatial robotic surgery datasets to enable this approach and validated it through extensive preclinical porcine in vivo studies in the context of automatic robotic pedicle screw placement surgery.

Our results demonstrate that \gls{rgb}-based localisation can be deployed effectively in robotic surgery, using spine surgery as an example, offering a compact and lightweight alternative to infrared tracking systems. In \secref{sec:results.proprioceptive_breathing_motion_estimation_in_the_robot_reference_frame}, we applied the proposed \glsentrylong{sdr} localisation strategy during a preclinically structured in vivo study, enabling automated pedicle screw placement with visual breathing compensation. The system achieved a 13-fold reduction in camera weight and a threefold reduction in size compared with conventional infrared setups, enabling bedside mounting with a direct line of sight to the surgical field. Importantly, this placement avoided obstruction by the surgeon or staff, ensuring uninterrupted visibility of the scene (\figref{fig:results.overview}A). As a result, marker visibility improved by $25\%$, providing more reliable coverage of the surgical site. Beyond improved visibility, the \gls{rgb}-based system reproduced the motions observed with infrared tracking (\figref{fig:results.breathing}) with submillimetre precision ($0.12/0.13/0.27\,\text{mm}$ vs. $0.05/0.06/0.04\,\text{mm}$ for \glsentrylong{ml}/\glsentrylong{cc}/\glsentrylong{ap}, respectively). Although slightly less accurate, this performance demonstrates that the substantial reduction in hardware burden was achieved without compromising clinically relevant tracking fidelity. 
Crucially, by directly observing the surgical site instead of a fiducial marker distant to it, the \gls{rgb}-based system revealed motions invisible to \glsentrylong{sota} clinical infrared tracking. The motion was captured $25\%$ closer to the drilling site ($8.1 \pm 0.1$ cm) and directly captured on the anatomy, providing breathing estimates with improved fidelity along the \glsentrylong{ml}, \glsentrylong{cc}, and \glsentrylong{ap} axes (\suptabref{tab:results.breathing_amplitudes}). Visually observed drilling-induced patient displacements were consistent with infrared tracking along the \glsentrylong{ml} axis, but \gls{rgb}-based tracking revealed a $250\%$ greater deformation in the drilling direction ($1.85 \pm 0.23$ mm vs. $0.53 \pm 0.06$ mm, \suptabref{tab:results.breathing_dynamics}). Along the \glsentrylong{cc} axis, the infrared marker even indicated motion opposite to the drilling direction ($0.33 \pm 0.17$ mm vs. $-0.64 \pm 0.20$ mm). Together, these findings suggest that conventional fiducial-based methods may substantially underestimate procedure-induced tissue and spine motion, with direct implications for trajectory accuracy.

Following the preclinical relevance studies, collected data was utilised to train a drape- and occlusion-invariant segmentation model. This segmentation model was subsequently applied to surgical robot localisation in a multi-robot setup. Using the proposed \glsentrylong{cso} algorithm followed by \glsentrylong{sdricp} refinement, localisation accuracy reached sub-percent deviations of less than $0.16\%$ ($0.39$ cm) at the robot base, and remained less than $0.65\%$ ($1.66$ cm) along the kinematic chain at a camera distance of $2.6\,\text{m}$ (\figref{fig:results.render_prior}, \figref{fig:results.localisation_errors_clinical}). This accuracy was observed despite severely different draping and surgical environment (\figref{fig:results.segmentations}). This level of precision is sufficient for safe collision avoidance in the operating theatre, particularly when combined with modest safety margins. The accuracy was further benchmarked against marker-based localisation methods. Notably, the proposed \glsentrylong{sdricp} approach outperformed the marker-based baseline under the common clinical limitation of unknown tool geometry, achieving lower median errors at the tool centre ([4.23, 3.13, 2.51, 1.88] mm vs. [2.87, 2.17, 1.78, 1.33] mm for 3, 6, 9, and 12 calibration robot configurations, respectively). These accuracies may already suffice for automated pedicle screw placement in the larger vertebrae (e.g., $13.61\,\text{mm}$ diameter at L5), but would require improvement for smaller ones (e.g., $5.09\,\text{mm}$ at T5)~\cite{pedicle_diameter_greek}. They are, however, more than adequate for tasks such as autonomous trocar insertion in laparoscopic surgery or for optimal patient positioning. When tool geometry was available, the marker-based approach achieved higher accuracy, outperforming \glsentrylong{sdricp} by $0.85\,\text{mm}$ (\figref{fig:results.localisation_errors_benchmark}). Importantly, when comparing draped and undraped marker-free localisation, the error for draped systems increased only within the submillimetre range ([1.04, 0.86, 0.57, 0.36] mm for \glsentrylong{sdricp}). This finding shows that the error introduced by draping can be effectively controlled, although localisation in surgical settings continues to pose challenges that will be discussed below.

\subsection*{Technical Challenges}
Drape-invariant segmentation remains challenging: the best \gls{iou} of $0.73 \pm 0.06$ observed in \suptabref{tab:results.iou} lies below what is commonly considered a good score ($>0.8$~\cite{segformer}), underscoring the difficulty of segmenting draped robots compared with related domains. By necessity, and given both the large volume of collected data and limited computational resources, segmentation and localisation were restricted to lower input resolutions of $512 \times 512$, which likely constrained achievable accuracy. Combining segmentation with localisation through an in-context prior (\figref{fig:results.render_prior}) mitigated the limitations of suboptimal \gls{iou} to a considerable extent, but at the cost of introducing strong dependence on the specific robot model in this supervised training scheme. This reliance limits deployability across surgical robotic platforms and would necessitate retraining or adaptation for each new system, which, however, is a tractable problem. Furthermore, although multi-robot segmentation was achieved via \glsentrylong{cmm} augmentation, localisation in this study was performed on a per-robot basis, with the number of robots in the scene assumed to be known a priori. While the proposed approach generalises across surgical tools and thereby offers maximum flexibility for surgeons, localisation errors increased on challenging clinical data toward the end-effector (\figref{fig:results.localisation_errors_clinical}). This effect was likely driven by tool-caused occlusions. Moreover, although our evaluation setting was intrinsically biased in favour of the marker-based method when tool geometry was available, the \glsentrylong{pnp}-based approach remains the clinical gold standard (\figref{fig:results.localisation_errors_benchmark}). Reliable convergence of the proposed gradient-based \glsentrylong{sdr} optimisation scheme still depends on careful initialisation. Although this was consistently achieved via the \glsentrylong{cso} algorithm, reliance on a two-stage process increases system complexity. Furthermore, the method assumes access to robot joint sensor readings, which may not be exposed by all commercial platforms. A unified optimisation approach across camera pose and joint parameters, without the need for initialisation, would be fundamentally more desirable to ensure robustness and simplicity.

\subsection*{Clinical Challenges}
Clinicians require the flexibility to rearrange equipment intraoperatively to respond to unanticipated complications. Like other localisation methods, the proposed \glsentrylong{sdr} approach assumes a static setup and requires recalibration after any rearrangement. Although \figref{fig:results.localisation_errors_benchmark} demonstrates that precise localisation under draping can be achieved with as few as six robot configurations , the reliance on manual recalibration remains a limitation, potentially reducing robustness to unforeseen intraoperative interruptions. While the proposed approach enables localisation via bedside-mounted stereo-\gls{rgb} cameras, new challenges, such as maintaining sterility in this setup (\figref{fig:results.overview}A), remain unaddressed. The demonstrated localisation accuracy may support tasks such as collision avoidance or patient positioning, but safety-critical autonomous actions, beyond those demonstrated during in vivo studies, will require further validation before clinical deployment.

\subsection*{Outlook}
Adapting \glsentrylong{sdr} to dynamic surgical scenes, where robots may be repositioned during procedures, will require only modest modifications. A more comprehensive dataset in terms of camera poses could be acquired through dynamic scene localisation, which would in turn support the development of more robust segmentation models and improve generalisability across surgical settings. The results of this research suggest the emergence of a positive feedback loop, in which improved localisation enables richer data acquisition, which in turn supports more accurate models. Localisation performance can therefore be expected to improve steadily moving forward. This trajectory may ultimately challenge the necessity of explicit calibration altogether, instead enabling an accurate, continuously updated representation of the surgical setup at any given time, akin to human perception of the scene, but with superhuman precision.

More ambitious research directions should aim to eliminate key assumptions made in this work, particularly the reliance on a known robot platform and the requirement to know the robot configurations a priori. Progress on the former may be enabled by recent advances in self-identifying robots~\cite{self_identifying_robots}, which leverage differentiable rendering techniques closely related to those presented here. Extending this paradigm could allow previously unknown surgical tools to be incorporated directly into the localisation process, potentially improving accuracy at the tool tip. Progress on both fronts would enable independence from specific robots and tools, paving the way for truly distributed surgical robotic systems.

In summary, this research delivers key advances in surgical scene understanding, highlighting that markers can be removed from the operating theatre. With minimal regulatory hurdles compared to fully autonomous systems, these methods lay the groundwork for first steps toward surgical autonomy and open the door to more advanced autonomous capabilities, especially during intelligent multi-robot interaction~\cite{roboballet}.

\section*{Methods}
\label{sec:methods}

\subsection*{Stereo AprilTag marker tracking}
\label{sec:methods.stereo_marker_tracking}
To estimate breathing motion and feedback the estimated breathing motion into the breathing motion compensation algorithm, we track AprilTag markers that are glued onto the anatomy (\figref{fig:results.overview}B). While AprilTag markers can be located monocularly, this often results in noisy pose estimates. To further establish a fair comparison with clinical stereo-based infrared marker tracking and to improve robustness, in this work we estimate marker positions through the bedside-mounted stereo RGB camera (\figref{fig:results.overview}A). This is simply achieved via triangulating the 3D position of the 2D left-right marker-centre correspondences via the readily available \glsentrylong{dlt} method~\cite{dlt} provided in OpenCV\footnote{\url{https://opencv.org/}}. While triangulating the AprilTag marker centres does only provide a position estimate and no orientation, knowledge of the robot's pose within the shared camera coordinate system does allow for breathing compensation regardless. To minimize image processing induced latency, AprilTags are only detected within a \glsentrylong{roi} of the \glsentrylong{hd} images at the surgical site. The camera intrinsics $\mathbf{K}$ are adjusted for this offset
\begin{equation}
    \mathbf{K}^{\prime} = \begin{bmatrix}
        f_x & 0   & c_x - t_x \\
        0   & f_y & c_y - t_y \\
        0   & 0   & 1
    \end{bmatrix}
\end{equation}
%
with the focal lengths $f_x$, $f_y$, the principal point $c_x$, $c_y$, and the top left crop offsets $t_x$, $t_y$. To further reduce latency, images are processed via intra-process communication over shared memory.

\subsection*{Stereo differentiable rendering and camera swarm optimisation}
\label{sec:methods.stereo_differentiable_rendering}
Depth information is readily available via a pre-trained depth estimation model in the deployed stereo-\gls{rgb} camera, but marker-free registration of the surgical robot into the camera coordinate system via point clouds proves infeasible due to the sterile drape and the consequential deviation of the surgical robot's mesh from them (\figref{fig:results.method_overview} point cloud). Instead we use an image-based differentiable rendering approach and exploit the translucent properties of sterile draping. Image-based differentiable rendering attempts to align a silhouette render of the surgical robot's virtual mesh with a segmentation of the real system, see \figref{fig:results.method_overview}. The rendering aspect of the approach can additionally be repurposed for creating ground-truth segmentations given a priori known accurate calibration (\figref{fig:results.transition}). Since \glsentrylong{mdr} may, however, not always align robustly along the optical axis, and a stereo-\gls{rgb} camera is readily available, we contribute a \glsentrylong{sdr} approach.

Let $\mathcal{V}_l = \left\{\mathbf{v}_i\in\mathbb{R}^4 \;\middle|\; i=0,\dots,N-1\right\}$ be the set of mesh vertices in homogeneous coordinates that comprise a robot link. In the robot's base frame, the configuration of each link is given through homogeneous transforms $\boldsymbol{\Theta}_l$ that are obtained via forward kinematics $\text{FK}(\mathbf{q}) = \left\{\boldsymbol{\Theta}_l\in\text{SE}(3) \;\middle|\; l=0,\dots\text{L}-1\right\}$ and the robot's 
joint configurations $\mathbf{q}\in\mathbb{R}^\text{D}$, with $\text{L}$ equalling the respective robot's number of links, $\text{D}$ being the number of \glsentrylong{dof}, and we compute forward kinematics using the \textsc{pytorch\_kinematics} library~\cite{pytorch_kinematics}. Together this yields the precomputed configured link vertices $\mathcal{V}^\prime_l = \boldsymbol{\Theta}_l\mathcal{V}_l$. To locate the robot with respect to the camera coordinate system in a stereo formulation now equates to jointly minimising a superimposed cost function $f$
\begin{equation}
    \argmin_{\boldsymbol{\Theta}_\text{left}} f(\mathbf{M}_\text{left}(\boldsymbol{\Theta}_\text{left}, \mathcal{V}^\prime_l), \mathbf{S}_\text{left}) + f(\mathbf{M}_\text{right}(\boldsymbol{\Theta}_\text{right}(\boldsymbol{\Theta}_\text{left}), \mathcal{V}^\prime_l), \mathbf{S}_\text{right})
    \label{eq:methods.objective}
\end{equation}
%
where $\mathbf{M}_{\text{left}/\text{right}}$ are binary silhouette renderings of the configured meshes in the left and right views, respectively (\figref{fig:results.method_overview} in green), and $\mathbf{S}_{\text{left}/\text{right}}$ are the robot's binary (drape-invariant) segmentations, displayed e.g. in \figref{fig:results.method_overview} and \figref{fig:results.segmentations} in blue. In this work, we choose the left camera frame $\boldsymbol{\Theta}_\text{left}$ as the reference coordinate system and parametrize the, through calibration known, and relatively fixed, right camera frame's pose $\boldsymbol{\Theta}_\text{right}$ therewith. Importantly, to guarantee that during optimisation the objective $\boldsymbol{\Theta}_\text{left}$ remains in $\text{SE}(3)$, one should optimise on continuous representations thereof, such as its Lie algebra $\mathfrak{se}(3)$. For availability in~\cite{pytorch_kinematics}, we optimise on a continuous 9D representation~\cite{rot6d}. Crucially, we optimise for the robot's pose in the camera frame, not vice versa, since optimising for the camera's pose would strongly entangle rotation and translation.

Omitting the right rendering for equivalence modulo constant left/right homogeneous transform, we next discuss how the rendering $\mathbf{M}_\text{left}$ is obtained. Initially, the configured vertices $\mathcal{V}^\prime_l$ are projected into clip space $\mathcal{C}^\prime_l$ through 
\begin{equation}
    \mathcal{C}^\prime_l=\mathbf{P}\boldsymbol{\Theta}_\text{left}\mathcal{V}^\prime_l
\end{equation}
%
where the perspective projection matrix $\mathbf{P}$ is given via
\begin{equation}
    \mathbf{P} = \begin{bmatrix}
        2f_x/w & 0      & 2c_x/w-1                                                & 0                                                       \\
        0      & 2f_y/h & 2c_y/h-1                                                & 0                                                       \\
        0      & 0      & (z_\text{max}+z_\text{min})/(z_\text{max}-z_\text{min}) & 2(z_\text{max}z_\text{min})/(z_\text{min}-z_\text{max}) \\
        0      & 0      & 1                                                       & 0
    \end{bmatrix}
    \label{eq:methods.perspective_projection}
\end{equation}
%
with $z_{\text{min}/\text{max}}$ being the minimum and maximum depths considered for rendering, respectively, and $w,h$ are the width and height of the rendered image. The clip space projected vertices $\mathcal{C}^\prime_l$ are then rendered following a standard rendering pipeline to obtain $\mathbf{M}_\text{left}$. We perform the rasterization and antialiasing steps in a differentiable manner using the \textsc{nvdiffrast} library~\cite{nvdiffrast}, and omit interpolation for the binary nature of the renders.

Many choices of the cost function $f$ in \eqref{eq:methods.objective} exist, e.g. through a simple \gls{mse}, \gls{iou} or Dice loss. In this work, we suggest two slightly more sophisticated formulations. The first cost function formulation emphasizes aligning the rendered robot's boundary over small segmentation inaccuracies, we thus express the objective as
\begin{equation}
    f(\mathbf{M}_\text{left},\mathbf{S}_\text{left}) = \frac{1}{hw}\|\mathbf{M}_\text{left} - d(\mathbf{S}_\text{left})\|^2_2
\end{equation}
%
which computes the \gls{mse} loss between the render $\mathbf{M}_\text{left}$ and the the segmentation's Euclidean distance transform $d$ as follows
\begin{equation}
    d(\mathbf{S}_\text{left}) =
    \begin{cases}
        \min\limits_{\mathbf{y} \in \partial \mathbf{S}_\text{left}} \|\mathbf{x} - \mathbf{y}\|_2, & \mathbf{x} \in \mathbf{S}_\text{left} \\
        0,                                                                                          & \mathbf{x} \notin \mathbf{S}_\text{left}
    \end{cases}
\end{equation}
%
%
for the segmentation's boundary $\partial \mathbf{S}_\text{left}$. Using this formulation, improves robustness to draping residuals at the base and tool adaptors at the end-effector, preventing precise binary segmentation of the robot itself through a foundation model. Second, with later access to a robust and occlusion-invariant segmentation model, we relax the assumption of imprecise segmentations and turn to a Dice loss variant
\begin{equation}
     f(\mathbf{M}_\text{left},\mathbf{S}_\text{left}) = 1 - \frac{2 \, \|\mathbf{M}_{\text{left}} \odot \exp(-d(\mathbf{1}-\mathbf{S}_{\text{left}}) / \sigma) \|_1}
{\|\mathbf{M}_{\text{left}}\|^2_2 + \|\exp(-d(\mathbf{1}-\mathbf{S}_{\text{left})} / \sigma)\|^2_2 + \epsilon}
\end{equation}
%
where $\odot$ denotes the Hadamard product, $d$ is again the Euclidean distance transform, $\sigma=2$ the exponential decay, and $\epsilon=1\mathrm{e}{-6}$ is a factor for numerical stability. The squared L2-norm in the denominator penalizes small misalignments, assuming precise final alignment. The exponential decay serves to increase the convergence basin and additionally smoothens the gradients at boundaries, whilst keeping the binary segmentation unchanged.

Either formulation requires a reasonable initialisation, shown e.g. in \figref{fig:results.render_prior}. In this work we base the camera pose initialisation off of the \glsentrylong{pso}~\cite{particle_swarm} algorithm. We simulate $N$ virtual cameras, each acting as a particle, thus referring to the algorithm as \glsentrylong{cso} algorithm. Each camera is initialized via random eye space coordinates. The eye is positioned randomly within a hollow sphere about the robot's base. The camera centres are initialized within a sphere about the robot's base. Since the number of virtual cameras $N$ is usually large (in the order of a couple thousand) and must fit into VRAM during rendering for effective parallelisation, we downscale the meshes by $95\%$ via quadric-based edge collapse simplification using the \textsc{fast\_simplification} library\footnote{\url{https://github.com/pyvista/fast-simplification}}.

\subsection*{Sensorimotor breathing compensation}
\label{sec:methods.sensorimotor_breathing_compensation}
We model the periodic \glsentrylong{ap} breathing motion $z(t)$ in the robot base frame via
\begin{equation} 
z(t) = a_0 + \sum_{n=1}^{3} a_n \cos(n\omega_0 t) +b_n \sin(n \omega_0 t)
\label{eq:methods.breathing_motion}
\end{equation}
%
where $a_0$, $a_n$, and $b_n$ are Fourier coefficients, $\omega_0$ is the breathing frequency, and $t$ is the current time. Here, visually observed respiratory motion $\Tilde{z}(t)$ is transformed into the robot reference frame~\eqref{eq:methods.objective} and is used to estimate the above parameters by minimising
\begin{equation}
\begin{aligned}
\min_{a_0, a_n, b_n, \omega_0} \quad & \int_{t-T_0}^{t}  \left( \Tilde{z}(t) - z(t) \right)^2 \,dt
\end{aligned}
\end{equation}
%
where $T_0=15\,\text{s} $ is the optimisation horizon. We cast the drill control as a quadratic problem
\begin{equation}
\begin{aligned}
\min_{\dot{\mathbf{q}}} \quad & \frac{1}{2} \dot{\mathbf{q}}^{T}\dot{\mathbf{q}}\\
\textrm{s.t.} \quad           & \mathbf{L}_\text{J} \leq \mathbf{J}\dot{\mathbf{q}} \leq \mathbf{U}_\text{J}
\end{aligned}
\label{eq:methods.control}
\end{equation}
%
where $\dot{\mathbf{q}}$ are the joint velocities, $\mathbf{L}_\text{J}/\mathbf{U}_\text{J}$ are the lower/upper bounds, $\mathbf{J}\in\mathbb{R}^{6\times \text{D}}$ is the Jacobian up to the drill tip, and D are again the \glsentrylong{dof}. We design bounds such that the drill advances with a desired force $f_d$ along a planned target trajectory under breathing compensation. The lower bound reads $\mathbf{L}_\text{J} = [\mathbf{v}; \boldsymbol{\omega}]$, where the angular velocity $\boldsymbol{\omega}$ enforces drill/trajectory alignment. The translational velocity $\mathbf{v}$ implements entry point following with compliance
\begin{equation}
    \mathbf{v} = \mathbf{R}[kx_\text{ep};ky_\text{ep};\dot{z}_\text{tip}-c(f_d-f_m)] + [0;0;\dot{z}(t)] 
\end{equation}
%
where $x/y_\text{ep}$ is the entry point in the tip frame, $k$ and $c$ are gains, $\dot{z}_\text{tip}$ is the drill tip velocity, and $f_m$ is the measured force. The kinematically obtained rotation $\mathbf{R}\in \text{SO}(3)$ rotates the expression into the robot base frame, and the breathing velocity $\dot{z}(t)$ is obtained through \eqref{eq:methods.breathing_motion}. The upper bounds equal the lower bounds, providing equality constraints. We implement~\eqref{eq:methods.control} within eTaSL~\cite{etasl}.

\subsection*{Segmentation models and training}
\label{sec:methods.segmentation_models_and_training}
\begin{figure}
    \centering
    \includegraphics[width=1.0\textwidth]{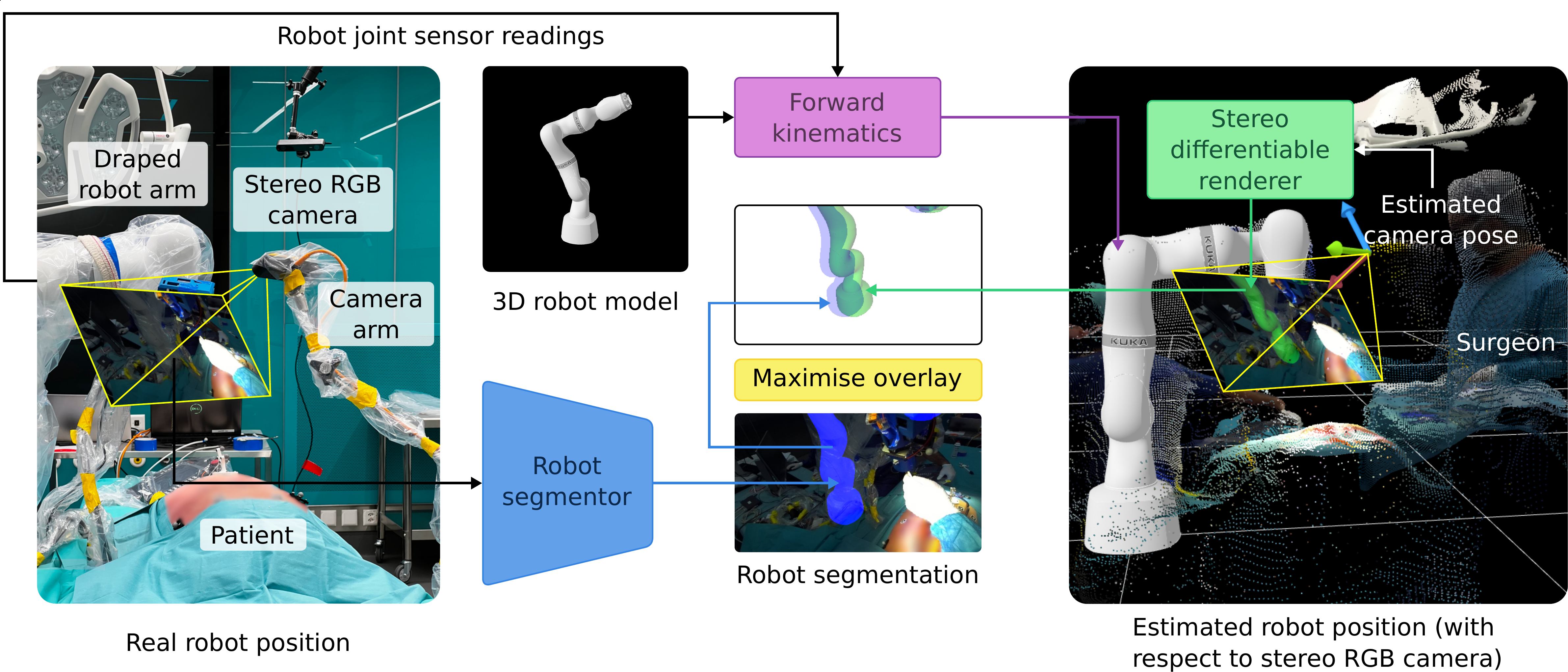}
    \caption{\textbf{\Glsentrylong{sdr}.} The proposed \glsentrylong{sdr} algorithm maximises the overlay of robot segmentations with the projection of a virtual robot model~\cite{lbr_stack}, thus yielding the robot location. 
    Initially, 
    SAM~2~\cite{sam2} was used to segment and subsequently locate the \textbf{undraped} surgical robot under preclinical workflow conditions. The robot was then \textbf{draped} in place and data got collected for \hyperref[sec:results.drape_and_occlusion_invariant_segmentation_of_surgical_robots]{\emph{drape- and occlusion-invariant segmentation of surgical robots}}, see \figref{fig:results.transition}. 
    For surgical \glsentrylong{ldn} mock spine surgery localisation benchmark, see \figref{fig:results.localisation_errors_benchmark}. The displayed data was obtained during the \glsentrylong{zh} human GAX cadaveric studies. The surgeon point cloud to the right was observed from the ceil-mounted stereo-\gls{rgb} camera and co-referenced to the surgical robot.
    }
    \label{fig:results.method_overview}
\end{figure}

\begin{figure}
    \centering
    \includegraphics[width=1.0\textwidth]{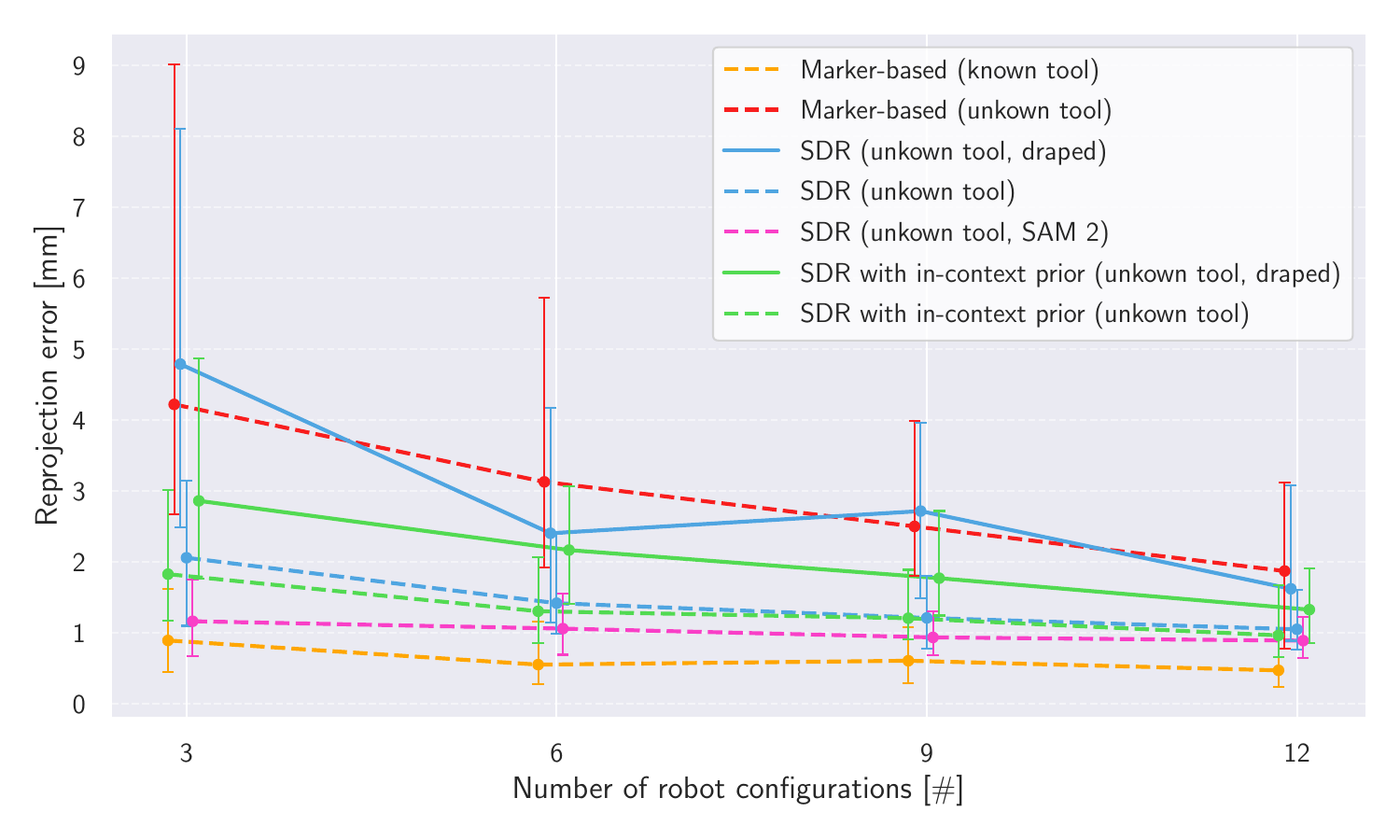}
    \caption{\textbf{\Glsentrylong{ldn} mock spine surgery localisation benchmark.}  
    Reprojection error medians, and $25\%$ (Q1) / $75\%$ (Q3) quantiles. For all \glsentryfull{sdr}-based methods (without in-context prior \figref{fig:results.method_overview}, with in-context prior \figref{fig:results.render_prior}), the reported error presents an upper bound, as described in \hyperref[sec:results.marker_free_surgical_robot_localisation_in_preclinical_workflow]{\emph{marker-free surgical robot localisation in preclinical workflow}}. Segmented via the MIT-B5-based three / four channel models (without / with in-context prior, \suptabref{tab:results.iou}) unless indicated as SAM~2. Dashed lines present undraped measurements, solid lines draped measurements, respectively. Benchmark data shown in \supfigref{fig:results.sie_benchmark_data}. For twelve robot configurations, \gls{sdr} (SAM~2) achieved a median error of $0.9\,/\,19\,\text{mm}$ (undraped / draped), \gls{sdricp} $0.97\,/\,1.33\,\text{mm}$ (confidently within $2\,\text{mm}$), and marker-based (known tool) $0.48\,\text{mm}$. For repeatability errors on \glsentrylong{par} multi-robot in vivo test dataset, refer \figref{fig:results.localisation_errors_clinical}.}
    \label{fig:results.localisation_errors_benchmark}
\end{figure}

Localisation via differentiable silhouette rendering, i.e. minimising \eqref{eq:methods.objective}, mandates a reference segmentation $\mathbf{S}$ (\figref{fig:results.method_overview}) that is grounded in real images. For undraped robots, we can segment the surgical robot using SAM~2~\cite{sam2}, however, this approach fails for draped systems. We thus require a segmentation model which remains invariant under draping and extracts the robot's contours regardless. In this work, we set to achieve drape invariance through supervised learning on ground-truth segmentations. Ground-truth segmentations are obtained via repurposing the presented rendering pipeline without gradient tracing and by kinematically tracking draped surgical robots throughout procedures (\figref{fig:results.transition} green overlay) using a pre-computed calibration.

For the segmentation pipeline we utilise a UNet-based~\cite{unet} model architecture with either transformer-based~\cite{segformer} or convolution-based~\cite{resnet} encoders through~\cite{pytorch_segmentation_models} (\suptabref{tab:results.iou}). For the decoder, we configure the UNet with five blocks of [256, 128, 64, 32, 16] feature layers, respectively, and finally predict single layer logits for the binary classification task. The logits are fed through a sigmoid function to obtain probabilities, and the loss $\mathcal{L}$ follows standard segmentation tasks as follows
\begin{equation}
    \mathcal{L}(\mathbf{M}, \mathbf{S}) = \omega_\text{focal}\mathcal {L}_\text{focal}(\alpha, \gamma, \mathbf{M}, \mathbf{S}) + \omega_\text{dice}\mathcal{L}_\text{dice}(\mathbf{M}, \mathbf{S})
\end{equation}
%
containing a focal $\mathcal{L}_\text{focal}$ and a dice $\mathcal{L}_\text{dice}$ component weighed by $\omega_\text{focal}=0.3$, $\omega_\text{dice}=0.7$. For the focal loss we use an exponent $\gamma=2$ and balance it through $\alpha=0.25$, as in the original paper~\cite{focal_loss}. Segmentation models are trained using the AdamW~\cite{adamw} optimiser with a learning rate of $1.5\mathrm{e}{-5}$ and a batch size of $B=18$ at an input resolution of $512\times512$ pixels.

To improve segmentation generalisation to various surgical theatre conditions, such as lighting, but also camera and robot placements, and varying number of surgical robots, we augment the ground-truth data during training (\figref{fig:results.transition}). Especially the varying number of surgical robots requires novel augmentations and particular consideration for the augmentation order. To this end, we propose a \glsentrylong{cmm} augmentation for simulating realistic multi-robot setups. The goal of the \glsentrylong{cmm} augmentation is to cut out robots via the binary render mask $\mathbf{M}$ and to insert it under geometric transforms into other images. This augmentation is constrained by robot visibility, since inserting partially visible robots would adversely affect ground-truth data. We thus distinguish between unconditional and conditional geometric transforms. Unconditional geometric transforms (e.g. horizontal flips) are applied in any case, and conditional geometric transforms (e.g. affine) are only applied given the robot is entirely visibility. Special care is also taken of the scaling for affine transformations. Since the training data is scaled for computational limitations from the common 16:9 to a 1:1 aspect ratio, as described above, but inference should equally function under scale-preserving cropping and scaling. To compensate for the modified aspect ratio, we scale randomly in the range [1.0, 1.8], which incorporates scale reversion. The \glsentrylong{cmm} operation then accumulates a batch $\mathcal{B}$ as follows
\begin{equation}
    \mathcal{B} = \left\{ (\mathbf{1}-\mathbf{M}^\prime_{j}) \odot \mathbf{I}^\prime_i
    + \mathbf{M}^\prime_{j} \odot \mathbf{I}^\prime_{j}
    \;\middle|\; i \in \{0,\dots,B-1\},\;
    j \in \operatorname{shuffle}\{0,\dots,B-1\} \right\}.
\end{equation}
%
where $\mathbf{1}$ denotes an all-ones matrix, $\mathbf{M}^\prime$ are the geometrically binary silhouette transformed renders, $\mathbf{I}^\prime$ are geometrically transformed images, $\odot$ again denotes the Hadamard product, and \text{shuffle} denotes a random permutation (uniform without replacement). Finally, we apply photometric transforms to the thus acquired batch $\mathcal{B}$. Following transforms are applied randomly: colour jitter, grayscale, box blur, contrast, salt and pepper noise, Gaussian noise, and channel dropout. Each photometric transform is applied with a $20\%$ probability and provided in the \textsc{kornia} library~\cite{kornia}.

\subsection*{Stereo differentiable rendering with in-context prior}
\label{sec:methods.stereo_differentiable_rendering_with_render_prior}
With availability to the full robot geometry and localisation estimates via the proposed \glsentrylong{sdr} and \glsentrylong{cso} algorithms, we exploit the opportunity to enhance robot segmentations. Incorporating localisation estimates for an in-context-guided segmentation process, we aim to alleviate some of the segmentation challenges imposed by the draping, as well as occlusions by tools, cables, tape, and surgical staff (\figref{fig:results.segmentations}A). Beyond segmenting from raw images we propose to guide the segmentation model via in-context priors~\cite{autocontext} (\figref{fig:results.render_prior}). In-context priors are realized by adding a fourth input channel to the segmentation model, appending the three channel \gls{rgb} input. When a localisation estimate is known, e.g. via \glsentrylong{cso} or \glsentrylong{sdr}, in-context priors are silhouette renders given the robot's configurations and the current best localisation estimate. When no localisation estimate is known yet, in-context priors are a zero input.

\begin{figure}
    \centering
    \includegraphics[width=1.0\textwidth]{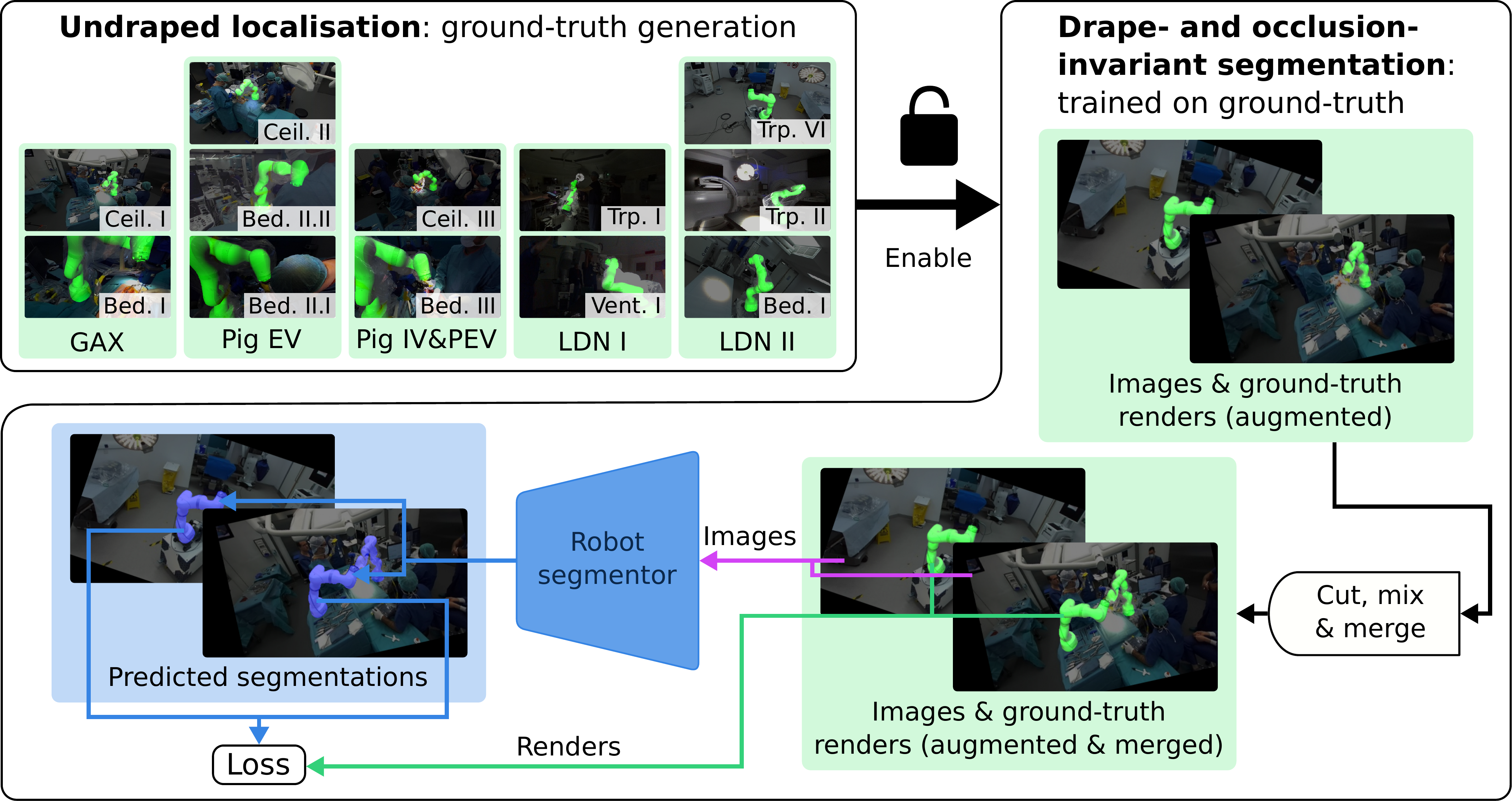}
    \caption{\textbf{Data collection and training scheme for drape- and occlusion-invariant segmentation of surgical robots.} (\textbf{Undraped localisation}) 
    Surgical robots were localised by the proposed \glsentrylong{sdr} algorithm 
    to kinematically generate ground-truth renders. Displayed samples from \glsentrylong{zh} human GAX cadaveric, porcine \glsentryfull{ev} / \glsentryfull{iv} / \glsentryfull{pev}, and \glsentryfull{ldn} datasets (details in \suptabref{tab:results.dataset}). (\textbf{Drape- and occlusion-invariant segmentation}) Transitioning the robot segmentor (also refer \figref{fig:results.method_overview}) to clinically realistic workflow conditions, i.e. \hyperref[sec:results.drape_and_occlusion_invariant_segmentation_of_surgical_robots]{\emph{drape- and occlusion-invariant segmentation of surgical robots}} was achieved on thus established ground-truth. The novel \glsentrylong{cmm} augmentation emulated multi-robot setups. For qualitative segmentation results refer \figref{fig:results.segmentations}.}
    \label{fig:results.transition}
\end{figure}

\begin{figure}
    \centering
    \includegraphics[width=\textwidth]{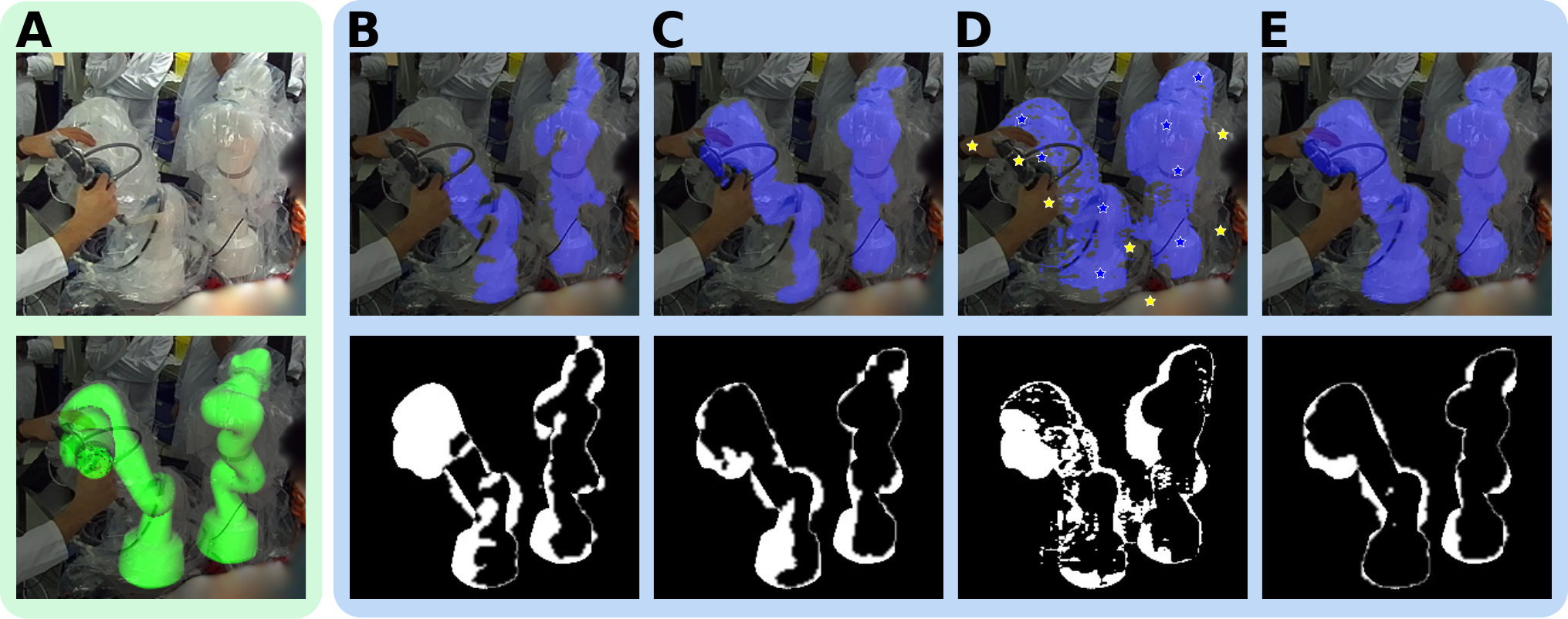}
    \caption{\textbf{Qualitative drape- and occlusion-invariant segmentation results on the \glsentrylong{par} multi-robot test dataset.} (\textbf{A}) Ground-truth renders as obtained by \hyperref[sec:results.marker_free_surgical_robot_localisation_in_preclinical_workflow]{\emph{marker-free surgical robot localisation in preclinical workflow}}. (\textbf{B}) MIT-B3 without \glsentrylong{cmm}. (\textbf{C}) MIT-B3 with \glsentrylong{cmm}. (\textbf{D}) SAM~2 foundation model. (\textbf{E}) MIT-B5 with \glsentrylong{cmm}. Top row shows segmentations, bottom row shows difference between the ground-truth renders and segmentations. The \glsentrylong{cmm} augmentation helped generalise \hyperref[sec:results.drape_and_occlusion_invariant_segmentation_of_surgical_robots]{\emph{drape- and occlusion-invariant segmentation of surgical robots}} to multi-robot setups. For quantitative metrics refer \suptabref{tab:results.iou}. Training scheme followed \figref{fig:results.transition}. Segmentations were consecutively used for \glsentrylong{sdr} (\figref{fig:results.method_overview}) and \glsentrylong{sdricp} (\figref{fig:results.render_prior}).}
    \label{fig:results.segmentations}
\end{figure}

\begin{figure}
    \centering
    \includegraphics[width=\textwidth]{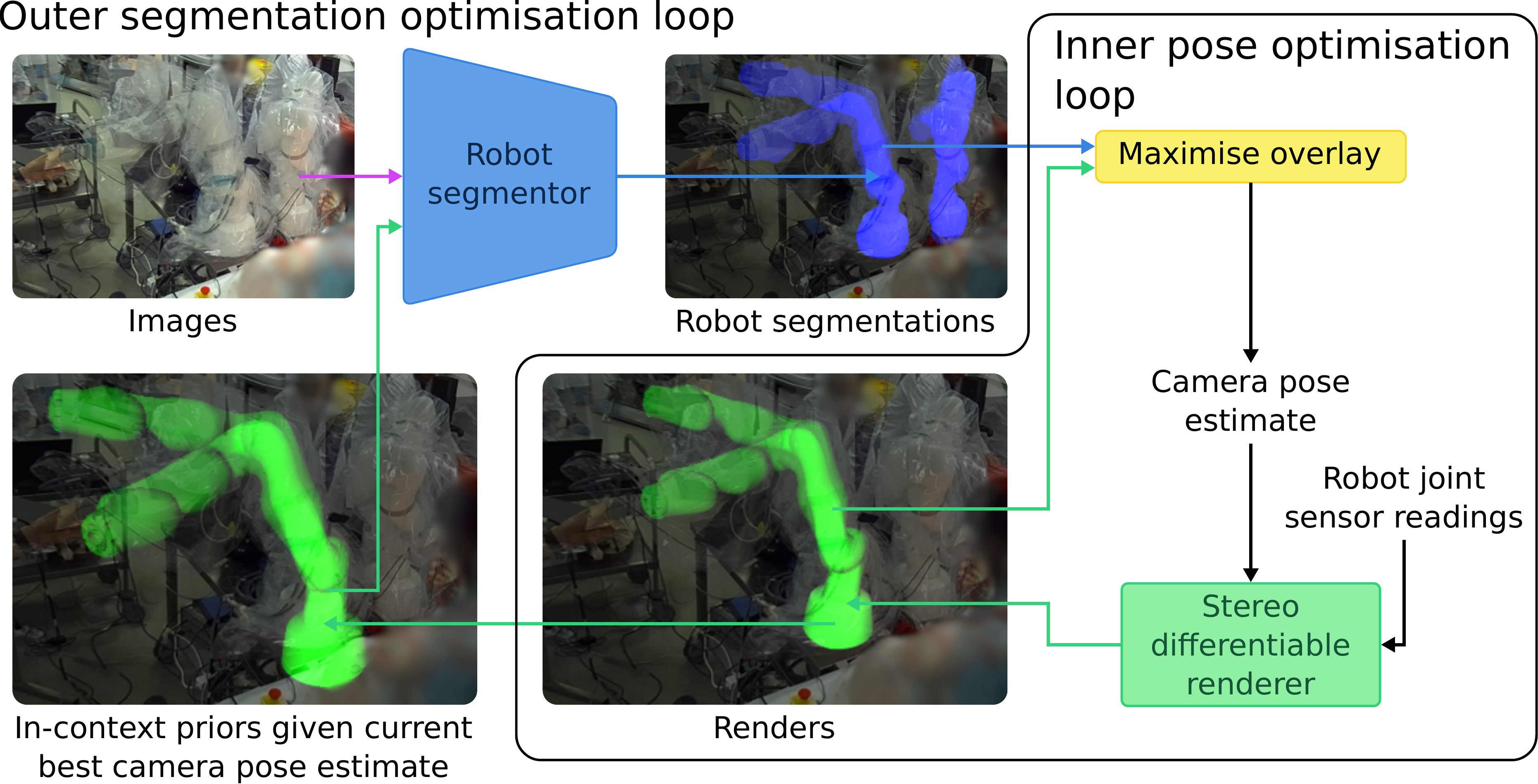}
    \caption{\textbf{\Glsentrylong{sdricp}.} Displayed are two overlaid configurations, but nine were used for localisation. The in-context prior shows the \glsentrylong{cso} initialization. Robot segmentations given the \glsentrylong{cso}-based in-context prior via the MIT-B5-based four in-channel model (\suptabref{tab:results.iou}, thee in-channel \figref{fig:results.segmentations}). The renders here show the converged result of the proposed \glsentrylong{sdricp} algorithm. Quantiative localisation accuracies are provided in \figref{fig:results.localisation_errors_benchmark} (\glsentrylong{ldn} mock spine surgery localisation benchmark) and \figref{fig:results.localisation_errors_clinical} (\glsentrylong{par} in vivo multi-robot test dataset).
    }
    \label{fig:results.render_prior}
\end{figure}

\begin{figure}
    \centering
    \includegraphics[width=\textwidth]{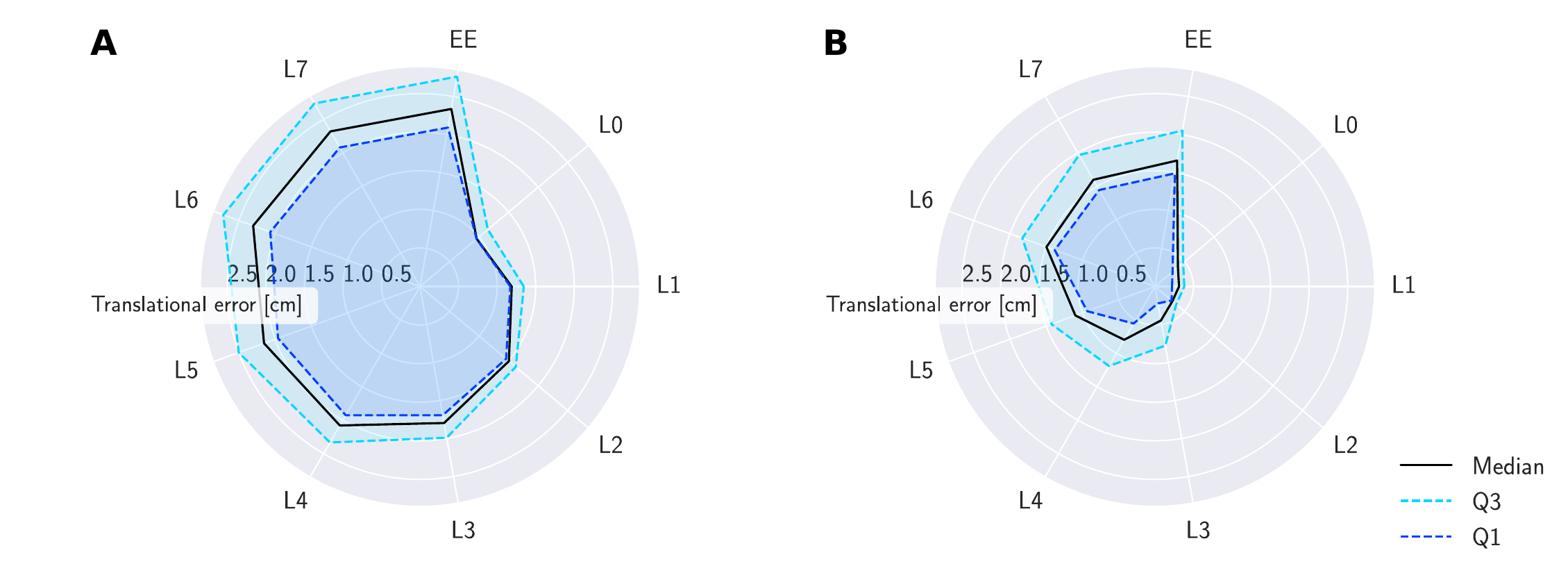}
    \caption{\textbf{Localisation repeatability errors on \glsentrylong{par} in-vivo spine surgery data across kinematic chain.} Undraped localisation obtained via \hyperref[sec:results.marker_free_surgical_robot_localisation_in_preclinical_workflow]{\emph{marker-free surgical robot localisation in preclinical workflow}} and draped localisation via \hyperref[sec:results.localising_draped_surgical_robots_inside_the_sterile_zone]{\emph{localising draped surgical robots inside the sterile zone}}, respectively. (\textbf{A}) \Glsentrylong{sdr}, refer \figref{fig:results.method_overview}. (\textbf{B}) \Glsentrylong{sdricp}, refer \figref{fig:results.render_prior}. Segmented via the MIT-B5-based three / four channel models (without / with in-context prior, \suptabref{tab:results.iou}). Kinematic chain from \glsentryfull{l0} to the \glsentryfull{ee}. For \glsentrylong{sdricp}, the largest observed median error (at the \glsentrylong{ee}) was $1.66\,\text{cm}$, i.e. $0.65\%$ at $2.6\,\text{m}$ camera distance. For \glsentrylong{ldn} mock spine surgery localisation benchmark refer \figref{fig:results.localisation_errors_benchmark}.
    }
    \label{fig:results.localisation_errors_clinical}
\end{figure}

To train a segmentation model with in-context prior incorporation given a localisation estimate, we render robot masks at training time under modified ground-truth camera poses \eqref{eq:methods.perspective_projection}. We therefore randomly sample translation deviations $\left\{\Delta\mathbf{x} \in \mathbb{R}^3 \;\middle|\; \Delta x_i \in [0, 5]\,\text{cm}, i=1,2,3\right\}$ and axis-angle rotation deviations on a sphere $\left\{\boldsymbol{\omega} \in \mathbb{R}^3 \;\middle|\; \|\boldsymbol{\omega}\|_2 = 5^\circ \right\}$ and use them to modify $\boldsymbol{\Theta}_\text{left}$. It is important to apply this rendering step in advance to the geometric augmentations of \figref{fig:results.transition}, and to scale the camera intrinsics for the 16:9 to 1:1 rescaling, as mentioned above.

During inference (\figref{fig:results.render_prior}), guiding the segmentation through in-context priors enables to incrementally improve segmentations by alternating between segmentation and \glsentrylong{sdr} steps. We call this process \glsentrylong{sdricp}. It encapsulates an inner pose optimisation loop within an outer segmentation optimisation loop, where the inner loop runs the suggested \glsentrylong{sdr} algorithm, and the outer loop updates the segmentation given the current best camera pose estimate.


\newpage

\begin{figure}
    \centering
    \includegraphics[width=\textwidth]{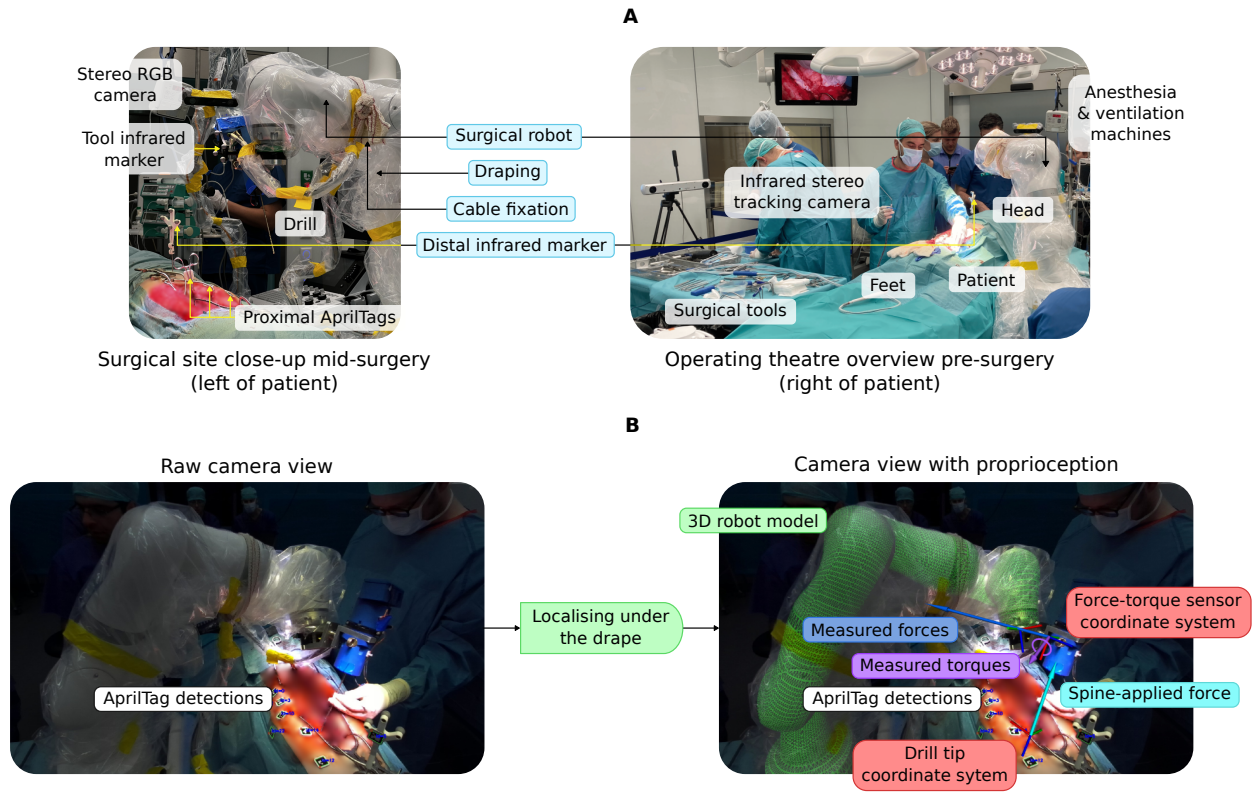}
    \caption{\textbf{Localising under the drape.} (\textbf{A}) Surgical setup during the \glsentrylong{zh} porcine in vivo study. 
    Following current clinical practise, a stereo infrared tracking camera was used to track spine breathing patterns at a distally screwed fiducial infrared marker (vertebrae T7). (\textbf{B}) View of the bedside-mounted stereo-\gls{rgb} camera during drilling at vertebrae T15 (left). The stereo-\gls{rgb} camera also tracked breathing patterns, however, at anatomy affixed AprilTags. Both cameras had exact knowledge of the surgical robot's location, through tool infrared markers, and via the proposed \glsentrylong{sdr}-based \hyperref[sec:results.marker_free_surgical_robot_localisation_in_preclinical_workflow]{\emph{marker-free surgical robot localisation in preclinical workflow}} (\figref{fig:results.method_overview}), respectively. This information allowed for referencing spine applied forces to the camera-observed breathing, i.e. \textbf{proprioception}, and was used for \hyperref[sec:results.breathing_compensated_drilling_in_robotic_porcine_in_vivo_spine_surgery]{\emph{breathing-compensated drilling in robotic porcine in vivo spine surgery}}, see \figref{fig:results.breathing} for detailed data.
    }
    \label{fig:results.overview}
\end{figure}

\begin{figure}
    \centering
    \includegraphics[width=\textwidth]{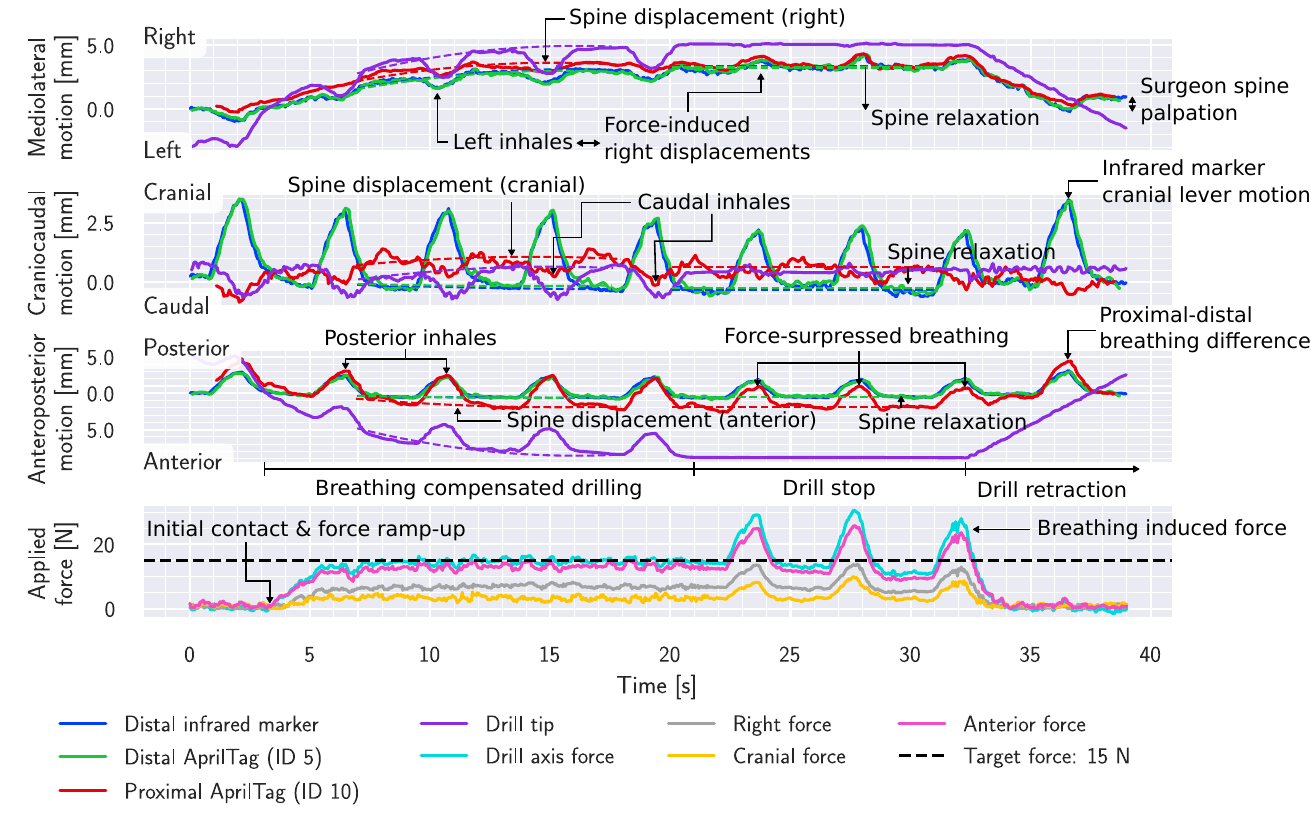}
    \caption{\textbf{Visually observed and kinematically tracked motions during drilling at vertebrae T15 (left).} Demonstrates \hyperref[sec:results.breathing_compensated_drilling_in_robotic_porcine_in_vivo_spine_surgery]{\emph{breathing-compensated drilling in robotic porcine in vivo spine surgery}} during the \glsentrylong{zh} in vivo study. Transitions from breathing compensated drilling (pre- and post-contact) to drill stop to drill retraction. Distal AprilTag  affixed onto fiducial distal infrared marker (\figref{fig:results.overview}B) and referenced via \hyperref[sec:results.proprioceptive_breathing_motion_estimation_in_the_robot_reference_frame]{\emph{proprioceptive breathing motion estimation in the robot reference frame}}. Quantitative measures are provided in \suptabref{tab:results.breathing_amplitudes} (isolated breathing amplitudes) and \suptabref{tab:results.breathing_dynamics} (dynamic spine and tissue displacements / relaxations).}
    \label{fig:results.breathing}
\end{figure}

\clearpage 

%
\bibliography{references} 
\bibliographystyle{sciencemag}

%
%
%
%
%
%


\section*{Acknowledgments}
\label{sec:acknowledgments}
\paragraph*{Funding:}
This work was supported by core funding from the Wellcome / EPSRC [WT203148 / Z / 16 / Z; NS / A000049 / 1], the European Union’s Horizon 2020 research and innovation programme under grant agreement No 101016985 (FAROS project), and EPSRC under the UK Government Guarantee Extension (EP / Y024281 / 1, VITRRO). This work has been supported by the OR-X -- a swiss national research infrastructure for translational surgery -- and associated funding by the University of Zurich and University Hospital Balgrist.
\paragraph*{Author contributions:}
M.H., T.V., and C.B. conceptualized methods and the experimental strategy. M.H., T.V., and C.E.M. developed core software components. M.H., A.D., R.L., A.H., E.V., and T.V. performed system integration. M.H. and T.V. collected data. M.H. and R.L. postprocessed data. M.H. developed and trained deep learning models. M.H., C.B., T.V., A.D., N.A.C., and C.E.M. prepared and authored the paper. M.H., C.B., T.V., and C.E.M. edited and revised the paper. N.A.C., F.C., F.T., and T.C. prepared preclinical studies. C.J.L., N.A.C., E.S, and M.F. performed surgeries. T.V., C.B., S.O., G.M., E.V., F.C., P.F., and M.F. managed the project and provided funding.



\paragraph*{Competing interests:}
S.O. and T.V. are co-founders and shareholders of Hypervision Surgical. 

\paragraph*{Data and materials availability:}
Data is made available upon request. Core registration code and segmentation models are made available at \url{https://github.com/lbr-stack/roboreg}. The robot driver and a 3D robot model are made available at \url{https://github.com/lbr-stack/lbr_fri_ros2_stack}.
\paragraph*{Ethics approvals}
Ethics approval has been granted for our \glsentrylong{zh} human ex vivo studies by the local ethics committee (BASEC 2021-01196). Ethics approval has been granted for our \glsentrylong{zh} porcine in vivo study by the Swiss Federal Food Safety and Veterinary Office under Ethical Application Number N°36440 Cantonal N° ZH003/2024. Ethics approval has been granted for our \glsentrylong{par} porcine in vivo study by the Comit\'{e} d’\'{e}thique Anses / ENVA / UPEC (registered with the French Comit\'{e} National de R\'{e}flexion Ethique sur l’Exp\'{e}rimentation Animale) under Agreement Number 94-046-02. 


\subsection*{Supplementary materials}
Materials and Methods\\
Figure S1\\
Table S1 to S4\\


\newpage


\renewcommand{\thefigure}{S\arabic{figure}}
\renewcommand{\thetable}{S\arabic{table}}
\renewcommand{\theequation}{S\arabic{equation}}
\renewcommand{\thepage}{S\arabic{page}}
\setcounter{figure}{0}
\setcounter{table}{0}
\setcounter{equation}{0}
\setcounter{page}{1} 


\begin{center}
\section*{Supplementary Materials for\\ \scititle}

Martin~Huber,
Nicola~A.~Cavalcanti,
Ayoob~Davoodi,
Ruixuan~Li,
Christopher~E.~Mower,
Fabio~Carrillo,
Christoph~J.~Laux,
Francois~Teyssere,
Thibault~Chandanson,
Antoine Harl\'{e},
Elie~Saghbiny,
Mazda Farshad,
Guillaume Morel,
Emmanuel~Vander~Poorten,
Philipp F\"urnstahl,
S\'{e}bastien~Ourselin,
Christos~Bergeles$^{\ast\dagger}$,
Tom~Vercauteren$^{\ast\dagger}$

\small$^\ast$Corresponding authors. Emails: christos.bergeles@kcl.ac.uk, tom.vercauteren@kcl.ac.uk\\
\small$^\dagger$These authors contributed equally to this work.
\end{center}

\subsubsection*{This PDF file includes:}
Materials and Methods\\
Supplementary Text\\
Figure S1\\
Table S1 to S4\\


\newpage


\subsection*{Materials and Methods}
\subsubsection*{Multi-centre spatial robotic surgery datasets}
\label{sec:supp.multi_centre_spatial_robotic_surgery_datasets}
For the \glsentrylong{par} dataset, we located two fully positioned surgical robots using a wall-mounted stereo-\gls{rgb} camera during a porcine \glsentrylong{iv} robotic pedicle screw placement procedure. Because this dataset includes two robots rather than one, it presents a more challenging localisation scenario compared to the other datasets, which each contain only a single robot.

For the \glsentrylong{zh} datasets (GAX, porcine \glsentrylong{ev} / \glsentrylong{iv} / \glsentrylong{pev}), we acquired data under realistic clinical conditions. In the \glsentrylong{pev} case, we humanely euthanised the animal under deep anaesthesia after more than five hours of surgery to minimise suffering in accordance with institutional and ethical guidelines, and then continued the procedure \glsentrylong{pev}. We recorded a single robot from both a ceiling-mounted and a bedside-mounted stereo-\gls{rgb} camera, which are visible in \figref{fig:results.method_overview} (left) during the GAX surgery. The camera positions varied across subjects (e.g., Bed. I and Bed. II) and within subjects (e.g., Bed. II.I and Bed. II.II). We assigned the major number to indicate correspondence between ceiling and bedside recordings, meaning that both stereo-\gls{rgb} cameras were co-referenced through the proposed \glsentrylong{sdr} localisation algorithm. Consequently, the datasets are spatially and temporally aligned and capture the surgical site from multiple perspectives. An exemplary alignment is shown in (\figref{fig:results.method_overview}, point cloud and render), where we fused corresponding ceiling and bedside observations into a \glsentrylong{vr} representation of the surgical theatre (\figref{fig:results.method_overview}, right). We also instrumented the surgical tool with a force/torque sensor (\figref{fig:results.overview}B) and recorded the measured forces for use in \glsentrylong{phri} with compliant admittance control as well as in \secref{sec:results.breathing_compensated_drilling_in_robotic_porcine_in_vivo_spine_surgery}. As shown in \figref{fig:results.overview}B, the \glsentrylong{zh} datasets include the weight-compensated force/torque signals in the sensor coordinate system and the corresponding transformed spine-applied forces. Furthermore, as analysed previously in \figref{fig:results.breathing}, the \glsentrylong{zh} datasets incorporate localisation of all AprilTags and the distal infrared marker for each procedure.

For the \glsentrylong{ldn} datasets (I and II), we acquired data to increase dataset diversity in robot joint state configurations and to collect additional robot-to-camera poses. While the \glsentrylong{zh} datasets provided rich sensory information and realistic clinical environments, including fully scrubbed surgical staff and variable lighting, their clinical nature constrained the range of robot joint states. To address this limitation, we recorded a single robot for each \glsentrylong{ldn} dataset, emphasising diverse robot configurations and robot-to-camera poses. Diverse robot configurations were achieved through collision-free path planning under environmental constraints~\cite{moveit}. For the \glsentrylong{ldn} I dataset, we recorded a surgical robot in an ophthalmic operating theatre using two cameras: a tripod-mounted stereo-\gls{rgb} camera and a monocular \gls{rgb}-depth camera positioned on a ventilation machine. The ophthalmic context was chosen because access to real surgical theatres is rare, and this opportunity provided valuable data under authentic clinical conditions. For the \glsentrylong{ldn} II dataset, we captured a single robot in a mock operating theatre using either a tripod-mounted or a bedside-mounted stereo-\gls{rgb} camera, resulting in a total of 11 distinct robot poses.

\subsection*{Supplementary Text}
\subsubsection*{In vivo breathing motion estimation fidelity}
The breathing motion estimation fidelity was evaluated at the fiducial infrared marker and the reference AprilTag (ID 5), refer \figref{fig:results.overview}B. Stereo tracking the AprilTags reduced the localisation noise of monocular tracking by $65\%$, from $0.38\,/\,0.31\,/\,0.80\,\text{mm}$ to $0.12\,/\,0.13\,/\,0.27\,\text{mm}$ (\glsentrylong{ml} / \glsentrylong{cc}/ \glsentrylong{ap}). While this represents a substantial improvement, the infrared tracking system remained slightly more reliable, with errors of $0.05\,/\,0.06\,/\,0.04\,\text{mm}$. Importantly, all measured errors were well within the submillimetre range and clinically acceptable for pedicle screw placement surgery.

To establish a physiological reference, an AprilTag was clamped directly onto the spinous process of the targeted vertebra T15 (\figref{fig:results.overview}A), and breathing amplitudes were compared with those measured by the tags used subsequently in \secref{sec:results.breathing_compensated_drilling_in_robotic_porcine_in_vivo_spine_surgery}. The corresponding amplitudes are summarised in \suptabref{tab:results.breathing_amplitudes}. The fiducial infrared marker exhibited lever-induced motion that was physiologically incorrect: along the \glsentrylong{cc} axis, it moved opposite to the breathing direction ($3.57 \pm 0.07\,\text{mm}$ cranially, red in \suptabref{tab:results.breathing_amplitudes}), and along the \glsentrylong{ml} axis it displayed $0.84 \pm 0.07\,\text{mm}$ motion to the left, where none was expected. In contrast, the proximal AprilTag (ID 10) showed physiologically representative amplitudes closer matching those of the clamped AprilTag. Furthermore, because the surgeon’s position intermittently obstructed the line of sight of the infrared tracking system, the fiducial marker was visible only $77.3\%$ of the time. By comparison, despite variable lighting conditions, the AprilTag was successfully triangulated $96.6\%$ of the time, yielding a $25\%$ relative increase in visibility.

\subsubsection*{Undraped localisation with drape-and occlusion-invariant segmentation model}
On the undraped mock spine surgery benchmark data, the proposed \glsentrylong{sdr} methods, without and with in-context prior, were found to perform similarly well, achieving median errors of [2.06, 1.42, 1.22, 1.06] mm and [1.83, 1.31, 1.21, 0.97] mm for [3, 6, 9, 12] robot configurations, respectively, where the in-context prior only demonstrated marginal accuracy improvements. In this undraped scenario, \glsentrylong{sdr} with the foundation segmentation model SAM~2 provided slightly more accurate localisations with [1.17, 1.06, 0.94, 0.9] mm median errors. We attribute the foundation model submillimetre localisation advantage to four circumstances. First and second, the SAM~2 foundation model was approximately three times the size (\suptabref{tab:results.iou}) and relied on user annotations (\figref{fig:results.segmentations}), thus possessed an inherent advantage. Third, with the MIT-B5-based segmentation models (three / four channels) trained on drape- and occlusion-invariance, occasional model uncertainties were observed, particularly around in the training datasets heavily occluded areas, such as the base and the \glsentrylong{ee}. Fourth, limited camera poses in the training datasets challenged deployment to this benchmark.

\newpage

\begin{figure}
    \centering
    \includegraphics[width=\textwidth]{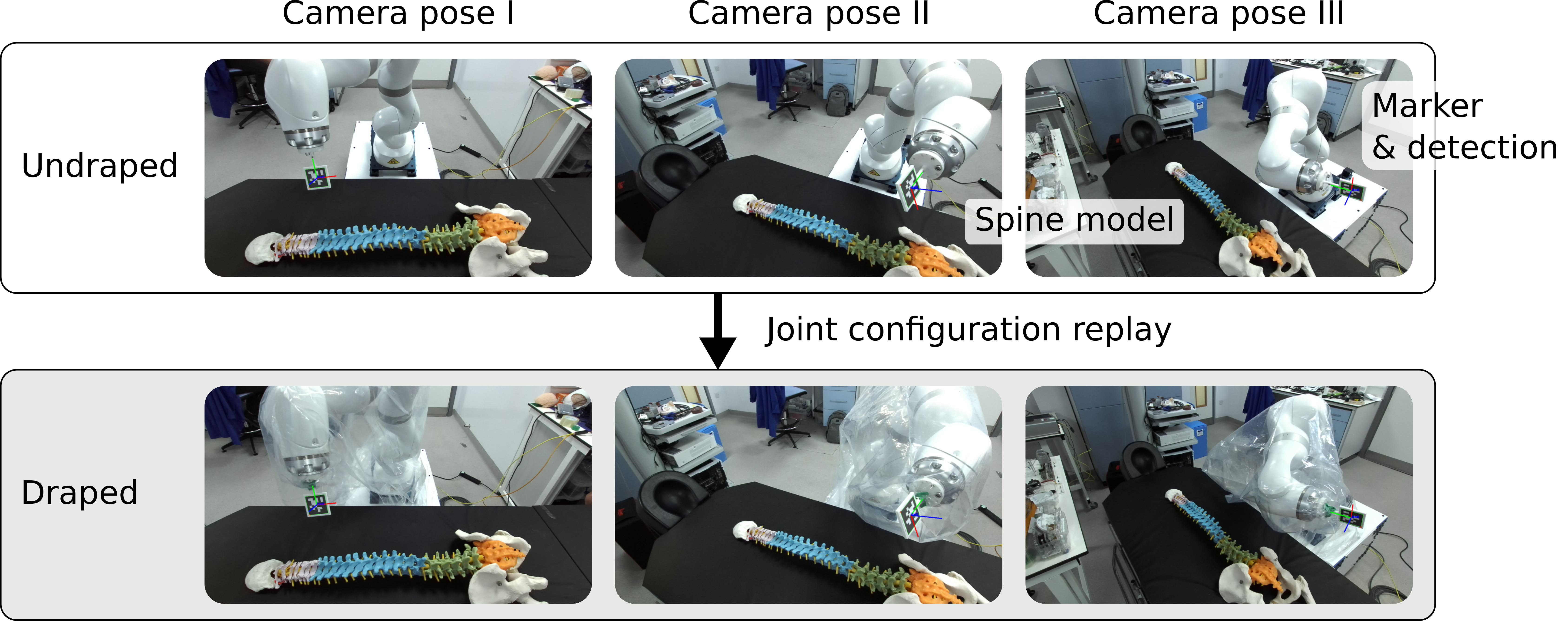}
    \caption{\textbf{Mock spine surgery tool centre localisation benchmark dataset.} Camera pose I: Left lateral thoracic level. Camera pose II: Left lateral sacral level. Camera pose III: Left lateral thigh level. Mimic \glsentrylong{zh} in vivo setup (\figref{fig:results.overview}).}
    \label{fig:results.sie_benchmark_data}
\end{figure}

\begin{table}
\caption{\textbf{Breathing amplitudes in isolation by location.} Amplitudes were acquired by \hyperref[sec:results.proprioceptive_breathing_motion_estimation_in_the_robot_reference_frame]{\emph{proprioceptive breathing motion estimation in the robot reference frame}} without external factors during the \glsentrylong{zh} porcine in vivo study. The clamped AprilTag was clamped immediately to the spinous process of vertebrae T15, see \figref{fig:results.overview}A. Red indicates amplitudes opposite to the physiologically plausible breathing direction: \glsentryfull{ml} negative = left, \glsentryfull{cc} negative = caudally, \glsentryfull{ap} positive = posteriorly.}
\label{tab:results.breathing_amplitudes}
\centering
\begin{tabular}{@{}lrrrr@{}}
    \toprule
    \multirow{2}{*}{Axis} & \multicolumn{4}{c}{Amplitude {[}mm{]}}                                                                      \\ \cmidrule(l){2-5} 
                          & Clamped AprilTag & Distal infrared marker        & Distal AprilTag              & Proximal AprilTag         \\ \midrule
    ML                    & $-0.04 \pm 0.07$ & $-0.84 \pm 0.07$              & $-0.74 \pm 0.09$             & $\mathbf{-0.14 \pm 0.16}$ \\
    CC                    & $-0.98 \pm 0.14$ & {\color{red}$3.57 \pm 0.07$}  & {\color{red}$3.64 \pm 0.08$} & $\mathbf{-1.29 \pm 0.26}$ \\
    AP                    & $3.87 \pm 0.16$  & $3.00 \pm 0.03$               & $3.11 \pm 0.07$              & $\mathbf{4.54 \pm 0.09}$  \\ \bottomrule
\end{tabular}
\end{table}

\begin{table}
\caption{\textbf{Quadratic fits to drilling-induced tissue displacements.} Displacement of surrounding tissue at proximal AprilTag (ID 10) and at distal infrared marker, including reference distal AprilTag (ID 5). Data acquired during \hyperref[sec:results.breathing_compensated_drilling_in_robotic_porcine_in_vivo_spine_surgery]{\emph{breathing compensated drilling in robotic in vivo spine surgery}} at vertebrae T15 (left). Red indicates displacements opposite to drilling direction: \glsentryfull{ml} positive = right, \glsentryfull{cc} positive = cranially, \glsentryfull{ap} negative = anteriorly. Also see data shown in \figref{fig:results.breathing}.}
\label{tab:results.breathing_dynamics}
\centering
\begin{tabular}{@{}clrrr@{}}
    \toprule
    Measurement                                                             & Location        & ML motion          & CC motion          & AP motion                       \\ \midrule
    \multirow{4}{*}{\shortstack{Displacement\\{[}mm{]}}}                    & Drill tip       & $ 5.045 \pm 0.033$ & $ 0.473 \pm 0.070$              & $-8.833 \pm 0.011$ \\ 
                                                                            & Dist. IR marker & $ 3.192 \pm 0.109$ & {\color{red}$-0.334 \pm 0.166$} & $-0.529 \pm 0.056$ \\
                                                                            & Dist. AprilTag  & $ 3.254 \pm 0.138$ & {\color{red}$-0.234 \pm 0.194$} & $-0.461 \pm 0.127$ \\
                                                                            & Prox. AprilTag  & $ 3.384 \pm 0.129$ & $ 0.643 \pm 0.204$              & $-1.850 \pm 0.225$ \\ \midrule
    \multirow{4}{*}{\shortstack{Displacement rate\\{[}mm/s{]}}}             & Drill tip       & $ 0.647 \pm 0.016$ & $ 0.134 \pm 0.007$              & $-1.136 \pm 0.028$ \\ 
                                                                            & Dist. IR marker & $ 0.465 \pm 0.010$ & {\color{red}$-0.037 \pm 0.007$} & $-0.088 \pm 0.005$ \\
                                                                            & Dist. AprilTag  & $ 0.464 \pm 0.014$ & {\color{red}$-0.036 \pm 0.012$} & $-0.111 \pm 0.012$ \\ 
                                                                            & Prox. AprilTag  & $ 0.482 \pm 0.024$ & $ 0.135 \pm 0.024$              & $-0.374 \pm 0.026$ \\ \midrule
    \multirow{4}{*}{\shortstack{Displacement\\rate change\\{[}mm/s$^2${]}}} & Drill tip       & $-0.026 \pm 0.001$ & $-0.008 \pm 0.001$              & $ 0.046 \pm 0.002$ \\ 
                                                                            & Dist. IR marker & $-0.019 \pm 0.001$ & {\color{red}$ 0.001 \pm 0.000$} & $ 0.004 \pm 0.000$ \\
                                                                            & Dist. AprilTag  & $-0.019 \pm 0.001$ & {\color{red}$ 0.002 \pm 0.001$} & $ 0.005 \pm 0.001$ \\ 
                                                                            & Prox. AprilTag  & $-0.022 \pm 0.002$ & $-0.007 \pm 0.002$              & $ 0.016 \pm 0.002$ \\ \bottomrule
\end{tabular}
\end{table}

\begin{table}
    \caption{\textbf{\glsentrylong{par}, \glsentrylong{zh}, and \glsentrylong{ldn} datasets.} Shown is the number of robots (Rob.), the type of camera (Cam.), i.e. \glsentryfull{m} and \glsentryfull{s}, use, i.e. test (Tst.), train (Trn.), and validation (Val.). \Glsentryfull{iv}, \glsentryfull{ev}, \glsentryfull{pev}. Further details are provided in \secref{sec:supp.multi_centre_spatial_robotic_surgery_datasets}. *\Glsentrylong{roi}: $1080\times1920$ pixels.}
    \label{tab:results.dataset}
    \centering
    \begin{tabular}{@{}lllrccrrc@{}}
    \toprule
    \multirow{2}{*}{Origin} & \multirow{2}{*}{Subject}                & \multirow{2}{*}{Location} & \multirow{2}{*}{ID} & \multirow{2}{*}{Rob./Cam.} & \multirow{2}{*}{Use} & \multicolumn{2}{c}{Frames}            & Resolution                 \\ 
                            &                                         &                           &                     &                            &                      & {[}$10^3${]}               & {[}Hz{]} & {[}Pixels{]}               \\ \midrule
    PAR                     & Porcine \acrshort{iv}                   & Wall I                    & 0                   & 2/S                        & Tst.                 &  41.5                      & 13       & $540\times960$             \\ \midrule
    \multirow{9}{*}{ZH}     & \multirow{2}{*}{GAX}                    & Ceil. I                   & 1                   & 1/S                        & Trn.                 & 243.8                      & 15       & $360\times640$             \\
                            &                                         & Bed. I                    & 2                   & 1/S                        & Trn.                 & 448.3                      & 28       & $\phantom{0}270\times480$* \\ \cmidrule(l){2-9} 
                            & \multirow{3}{*}{Porcine \acrshort{ev}}  & Ceil. II                  & 3                   & 1/S                        & Val.                 &  61.2                      & 6        & $1080\times1920$           \\
                            &                                         & Bed. II.I                 & 4                   & 1/S                        & Val.                 &  39.8                      & 18       & $1080\times1920$           \\
                            &                                         & Bed. II.II                & 5                   & 1/S                        & Val.                 &  41.2                      & 18       & $1080\times1920$           \\ \cmidrule(l){2-9} 
                            & \multirow{2}{*}{Porcine \acrshort{iv}}  & Ceil. III                 & 6                   & 1/S                        & Trn.                 &  61.1                      & 5        & $1080\times1920$           \\
                            &                                         & Bed. III                  & 7                   & 1/S                        & Trn.                 & 165.5                      & 14       & $1080\times1920$           \\ \cmidrule(l){2-9} 
                            & \multirow{2}{*}{Porcine \acrshort{pev}} & Ceil. III                 & 8                   & 1/S                        & Trn.                 &  15.5                      & 5        & $1080\times1920$           \\
                            &                                         & Bed. III                  & 9                   & 1/S                        & Trn.                 &  37.0                      & 12       & $1080\times1920$           \\ \midrule
    \multirow{2}{*}{LDN I}  & \multirow{2}{*}{N/A}                    & Vent. I                   & 10                  & 1/M                        & Trn.                 &   5.7                      & 19       & $\phantom{0}720\times1280$ \\
                            &                                         & Trp. I                    & 11                  & 1/S                        & Trn.                 &   5.0                      & 8       & $\phantom{0}720\times1280$ \\ \midrule 
    \multirow{6}{*}{LDN II} & \multirow{2}{*}{N/A}                    & Trp. I                    & 12                  & 1/S                        & Val.                 &  28.9                      & 15       & $540\times960$             \\
                            &                                         & Trp. II-III               & 13-14               & 1/S                        & Trn.                 &  90.0                      & 15       & $540\times960$             \\ \cmidrule(l){2-9} 
                            & \multirow{4}{*}{N/A}                    & Bed. I-III                & 15-17               & 1/S                        & Trn.                 &  60.4                      & 15       & $540\times960$             \\
                            &                                         & Bed. IV                   & 18                  & 1/S                        & Val.                 &  12.4                      & 15       & $540\times960$             \\ 
                            &                                         & Trp. IV-VI                & 19-21               & 1/S                        & Trn.                 &  70.1                      & 15       & $540\times960$             \\
                            &                                         & Trp. VII                  & 22                  & 1/S                        & Val.                 &   6.0                      & 15       & $540\times960$             \\ \bottomrule
\end{tabular}
\end{table}

\begin{table}
\caption{\textbf{Quantitative drape- and occlusion-invariant segmentation results on the \glsentrylong{par} multi-robot test dataset.} The \glsentryfull{cmm} augmentation helped generalise \hyperref[sec:results.drape_and_occlusion_invariant_segmentation_of_surgical_robots]{\emph{drape- and occlusion-invariant segmentation of surgical robots}} to multi-robot setups. SAM~2 achieved $0.60 \pm 0.03$ \glsentryfull{iou} with $224.4\mathrm{e}{6}$ parameters. For qualitative segmentation results refer \figref{fig:results.segmentations}. Training scheme followed \figref{fig:results.transition}. The MIT-B3-based model with \gls{cmm} augmentation is distributed as  roboseg-v0-medium. The MIT-B5-based model with three in-channels is distributed as roboseg-v0-large.}
\label{tab:results.iou}
\centering
\begin{tabular}{@{}llccccl@{}}
\toprule
Architecture           & Encoder    & Param. {[}$1\mathrm{e}{6}${]} & IoU {[}a.u.{]}           & CMM    & In-channels & Dataset                 \\ \midrule
\multirow{5}{*}{U-Net} & ResNet-101 & $51.5$              & $0.35 \pm 0.16$          & \xmark & 3        & \multirow{3}{*}{medium} \\
                       & MIT-B3     & $47.4$              & $0.64 \pm 0.10$          & \xmark & 3        &                         \\
                       & MIT-B3     & $47.4$              & $\mathbf{0.69 \pm 0.07}$ & \cmark & 3        &                         \\ \cmidrule(l){2-7}
                       & MIT-B5     & $84.7$              & $\mathbf{0.73 \pm 0.06}$ & \cmark & 3        & \multirow{2}{*}{large}  \\
                       & MIT-B5     & $84.7$              & $\mathbf{0.73 \pm 0.06}$ & \cmark & 4        &                         \\ \bottomrule
\end{tabular}
\end{table}


\clearpage 





\end{document}